\documentclass[manuscript]{acmart}

\AtBeginDocument{%
  \providecommand\BibTeX{{%
    \normalfont B\kern-0.5em{\scshape i\kern-0.25em b}\kern-0.8em\TeX}}}

\setcopyright{acmcopyright}
\copyrightyear{2023}
\acmYear{2023}

\usepackage{amsmath,amsfonts,bm}

\def\eqref#1{equation~\ref{#1}}

\def\1{\bm{1}}

\DeclareMathAlphabet{\mathsfit}{\encodingdefault}{\sfdefault}{m}{sl}
\SetMathAlphabet{\mathsfit}{bold}{\encodingdefault}{\sfdefault}{bx}{n}

\newcommand{\E}{\mathbb{E}}

\usepackage{color}
\usepackage{graphics}
\usepackage{hyperref}
\usepackage{url}
\usepackage{subfigure, multirow, array}
\usepackage{xcolor}
\usepackage[linesnumbered,ruled,vlined]{algorithm2e}
\usepackage{booktabs}
\usepackage[normalem]{ulem}
\useunder{\uline}{\ul}{}
\usepackage{wrapfig}
\usepackage{enumitem}
\newcommand{\adatime}{\textsc{AdaTime}\xspace}
\usepackage{caption} 
\usepackage{nicematrix}
\usepackage{color,array}
\usepackage{booktabs}
\usepackage{amsmath}
\usepackage{bbm}
\usepackage{bm}
\makeatletter

\def\zapcolorreset{\let\reset@color\relax\ignorespaces}
\def\colorrows#1{\noalign{\aftergroup\zapcolorreset#1}\ignorespaces}

\begin{document}

\title{ADATIME: A Benchmarking Suite for Domain Adaptation on Time Series Data}

\author{Mohamed Ragab}
\authornote{Both authors contributed equally to this research.}
\affiliation{%
\institution{Institute for Infocomm Research and Centre for Frontier AI Research, A*STAR}
\streetaddress{1 Fusionopolis Way}
\country{Singapore}
\postcode{138632}}
\email{mohamedr002@e.ntu.edu.sg}
\orcid{0000-0002-2138-4395}

\author{Emadeldeen Eldele}
\authornotemark[1]
\email{emad0002@ntu.edu.sg}
\orcid{0000-0002-9282-0991}
\affiliation{
  \institution{Nanyang Technological University}
  \streetaddress{50 Nanyang Ave}
  \country{Singapore}
  \postcode{639798}
}

\author{Wee Ling Tan}
\affiliation{%
  \institution{Institute for Infocomm Research, A*STAR}
  \streetaddress{1 Fusionopolis Way}
  \country{Singapore}
  \postcode{138632}}
\email{weeling.tan@eng.ox.ac.uk}

\author{Chuan-Sheng Foo}
\affiliation{%
  \institution{Institute for Infocomm Research and Centre for Frontier AI Research, A*STAR}
  \streetaddress{1 Fusionopolis Way}
  \country{Singapore}
  \postcode{138632}}
\email{foo_chuan_sheng@i2r.a-star.edu.sg}

\author{Zhenghua Chen}
\authornote{Corresponding Author}
\affiliation{%
  \institution{Institute for Infocomm Research and Centre for Frontier AI Research, A*STAR}
  \streetaddress{1 Fusionopolis Way}
  \country{Singapore}
  \postcode{138632}}
\email{chen0832@e.ntu.edu.sg}

\author{Min Wu}
\affiliation{%
  \institution{Institute for Infocomm Research, A*STAR}
 \streetaddress{1 Fusionopolis Way}
  \country{Singapore}
  \postcode{138632}}
\email{wumin@i2r.a-star.edu.sg}

\author{Chee-Keong Kwoh}
\affiliation{
  \institution{Nanyang Technological University}
  \streetaddress{50 Nanyang Ave}
  \country{Singapore}
  \postcode{639798}}
\email{asckkwoh@ntu.edu.sg}

\author{Xiaoli Li}
\affiliation{%
  \institution{Institute for Infocomm Research and Centre for Frontier AI Research, A*STAR}
 \streetaddress{1 Fusionopolis Way}
   \country{Singapore}
  \postcode{138632}}
\affiliation{
  \institution{Nanyang Technological University}
  \streetaddress{50 Nanyang Ave}
  \country{Singapore}
  \postcode{639798}}
\email{xlli@i2r.a-star.edu.sg}

\renewcommand{\shortauthors}{Mohamed Ragab and Emadeldeen Eldele, et al.}

\begin{abstract}

Unsupervised domain adaptation methods aim to generalize well on unlabeled test data that may have a different (shifted) distribution from the training data. Such methods are typically developed on image data, and their application to time series data is less explored. Existing works on time series domain adaptation suffer from inconsistencies in evaluation schemes, datasets, and backbone neural network architectures. Moreover, labeled target data are often used for model selection, which violates the fundamental assumption of unsupervised domain adaptation. To address these issues, we develop a benchmarking evaluation suite (\adatime) to systematically and fairly evaluate different domain adaptation methods on time series data. Specifically, we standardize the backbone neural network architectures and benchmarking datasets, while also exploring more realistic model selection approaches that can work with no labeled data or just a few labeled samples. Our evaluation includes adapting state-of-the-art visual domain adaptation methods to time series data as well as the recent methods specifically developed for time series data. We conduct extensive experiments to evaluate 11 state-of-the-art methods on five representative datasets spanning 50 cross-domain scenarios. Our results suggest that with careful selection of hyper-parameters, visual domain adaptation methods are competitive with methods proposed for time series domain adaptation. In addition, we find that hyper-parameters could be selected based on realistic model selection approaches. Our work unveils practical insights for applying domain adaptation methods on time series data and builds a solid foundation for future works in the field. The code is available at \href{https://github.com/emadeldeen24/AdaTime}{github.com/emadeldeen24/AdaTime}.
\end{abstract}

\maketitle

\section{Introduction}

Time series classification problem is predominant in many real-world applications, including healthcare and manufacturing. 
Recently, deep learning has gained more attention in time series classification tasks. It aims to learn the temporal dynamics in the complex underlying data patterns, assuming access to a vast amount of labeled data \cite{fawaz2019deep,lines2018time}. Yet, annotating time series data can be challenging and burdensome due to its complex nature that requires expert domain knowledge \cite{har_sys,eldele2021time,li2021modeling,li2021domain,sharma2021quick}. One way to reduce the labeling burden is to leverage annotated data (e.g., synthetic or public data) from a relevant domain (i.e., source domain) for the model's training while testing the model on the domain of interest (i.e., target domain). However, the source and target domains may have distinct distributions, resulting in a significant domain shift that hinders the model performance on the target domain. Such a problem commonly exists in many time series applications, including human activity recognition~\cite{har_sys,codats} and sleep stage classification (SSC) tasks~\cite{eldele2021adversarial}. For instance, a model can be trained to identify the activity of one subject (i.e., source domain) and tested on data from another subject (i.e., target domain), leading to poor performance caused by the domain shift problem.

Unsupervised Domain Adaptation (UDA) aims to transfer knowledge learned from a labeled source domain to an unseen target domain, tackling the domain shift problem. Much literature has been proposed for UDA on visual applications \cite{tl_survey,visual_da,wu2022multiple}. One prevailing paradigm aims to minimize statistical distribution measures to mitigate the distribution shift problem between the source and target domains \cite{deepcoral,HoMM,ddc,dsan,zhong2009cross}. Another promising paradigm that has recently emerged leverages adversarial training techniques to mitigate the domain gap \cite{long2015learning,ganin2016domain,CDAN,slarda}, inspired by generative adversarial networks \cite{goodfellow_gan}.

Recently, more attention has been paid to Time Series UDA (TS-UDA) \cite{har_sys,codats,dskn}. However, previous work on TS-UDA methods suffers from the following limitations. First, most of the existing algorithms specialize to particular application or domain \cite{har_sys,ragab2020adversarial,Li2020DomainAF}. Thus, there is an apparent shortage of baseline methods when applying domain adaptation to time series data. Second, existing TS-UDA works lack consistent evaluation schemes including benchmark datasets, preprocessing, and backbone networks. For instance, methods using recurrent neural networks as a backbone network \cite{vrada} have been compared against methods with convolutional neural network-based backbone networks \cite{codats}. In addition to differences in backbones, training procedures can also vary between different algorithms in terms of the number of epochs, weight decay, and learning rate schedulers \cite{vrada,rdann}, which results in an inconsistent evaluation of new algorithms. Last, most of the existing TS-UDA approaches often utilize labeled data from the target domain for model selection, violating the unsupervised assumption of UDA \cite{codats,dskn}, and providing an over-optimistic view of their real-world performance. The aforementioned issues can contribute to the performance, and the performance gain is mistakenly attributed to the proposed UDA method.

\begin{figure*}
    \centering
    \resizebox{\textwidth}{!}{
    \includegraphics{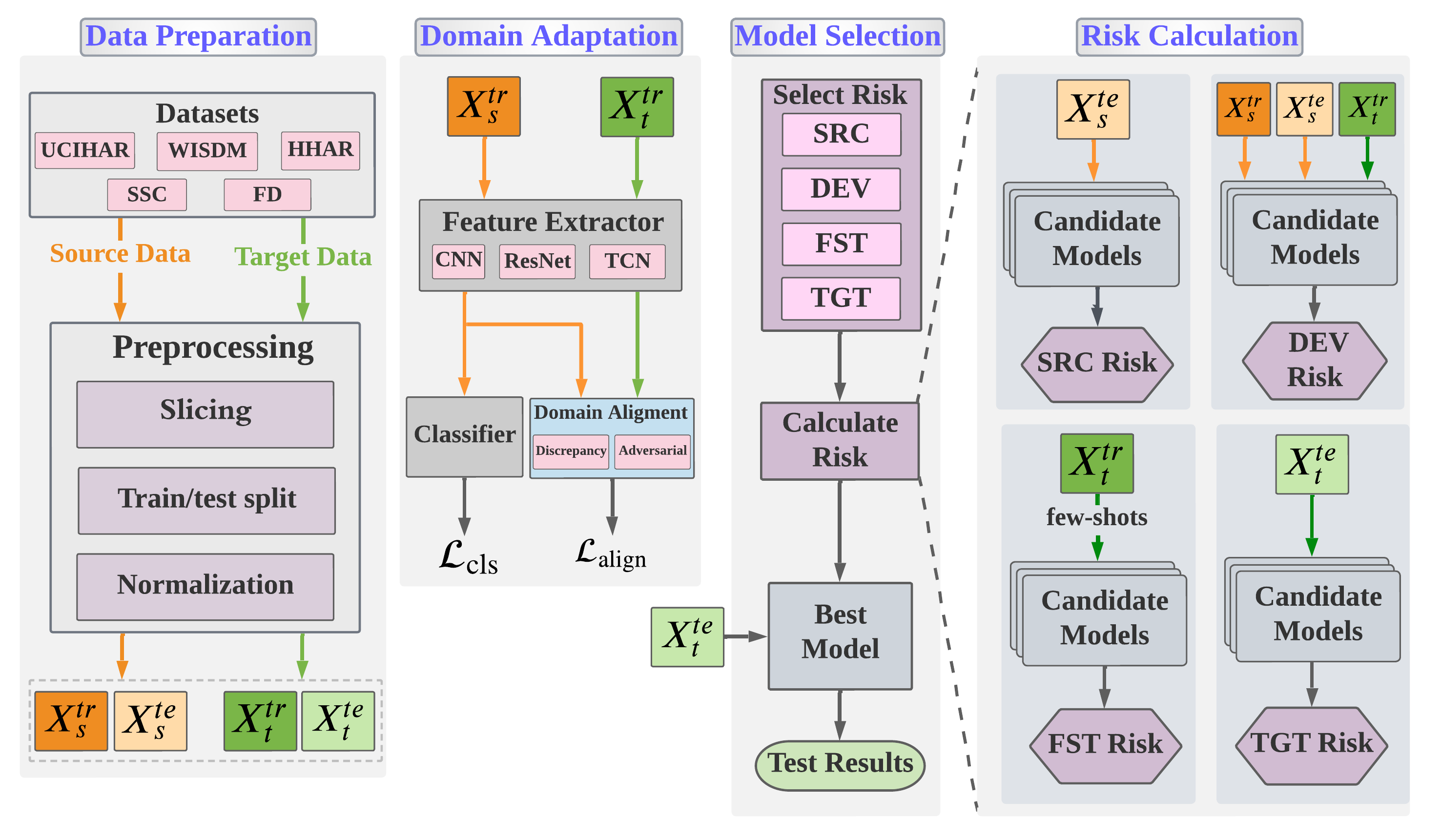}}
    \caption{Our benchmarking suite \adatime consists of three main steps: Data Preparation, Domain Adaptation, and Model Selection. We first prepare the train and test data for both source and target domains (i.e.,  $X^{tr}_s, X^{te}_s, X^{tr}_t, X^{te}_t$). Then the training sets of source and target domains are passed through the backbone network to extract the corresponding features. The domain alignment algorithm being evaluated is then used to address the distribution shift between the two domains. Last, given a specific risk type, we calculate the risk value for all the candidate models and then select the hyper-parameters of the one achieving the lowest risk. The selected model is lastly used for reporting the test results given the target domain test set (best viewed in colors).}
    \label{fig:framework}
\end{figure*}

In this work, we develop a systematic evaluation suite (\adatime) to tackle the aforementioned obstacles and remove all extraneous factors to ensure a fair evaluation of different UDA algorithms on time series data. In \adatime, we include current TS-UDA methods and re-implemented various state-of-the-art visual UDA methods that can be adapted to time series data. To ensure a fair evaluation of these methods, we standardize backbone networks and training procedures, data preparation, and preprocessing to address the inconsistency in previous evaluation schemes. Then, to select the model hyper-parameters, we explore more realistic model selection strategies that do not require target labels. Therefore, the main contributions of this paper can be summarized as follows: 

\begin{itemize}
    \item We systematically and fairly evaluate existing UDA methods on time series data. To the best of our knowledge, this is the first work to benchmark different UDA methods on time series data.
    
    \item We develop a benchmarking evaluation suite (\adatime) that uses a standardized evaluation scheme and realistic model selection techniques to evaluate different UDA methods on time series data.
    
    \item We evaluate 11 state-of-the-art UDA methods on five representative time series datasets spanning 55 cross-domain scenarios, and present comprehensive conclusions and recommendations for the TS-UDA problem. These evaluation results and analysis can provide a systematic guideline for future research on TS-UDA. 
\end{itemize}

The following sections are organized as follows. In Section \ref{sec:DA_definition}, we define the unsupervised domain adaptation problem and how adaptation is generally achieved. Section \ref{sec:adatime} describes the main components of our \adatime suite such as benchmarking datasets, unified backbone networks, adapted UDA algorithms, model selection approaches, and unified evaluation schemes. Section \ref{sec:results} shows the evaluation results and discusses the main findings of our experiments. Section \ref{sec:conclusion} presents the main conclusions and recommendations.

\section{Domain Adaptation}
\label{sec:DA_definition}
\subsection{Problem Formulation}
We start by defining the unsupervised domain adaptation problem. 
We assume access to labeled data from a source domain $\mathcal{X}_s = \{(x^i_s, y^i_s)\}_{i=1}^{N_s}$ that represents univariate or multivariate time series data, and unlabeled data from a target domain $\mathcal{X}_t = \{(x^j_t)\}_{j=1}^{N_t}$,  where $N_s$ and $N_t$ denote the number of samples for $\mathcal{X}_s$ and $\mathcal{X}_t$ respectively. Here we focus on classification and assume that both domains share the same label space $Y=\{1,2, \dots K\}$, where $K$ is the number of classes. 
Upon preprocessing, the source domain is split into a training set $X_s^{tr}$ with $N_s^{tr}$ samples, and a test set $X_s^{te}$ with $N_s^{te}$ samples. Similarly, the target domain is split into a training set $X_t^{tr}$ with $N_t^{tr}$ samples, and a test set $X_t^{te}$ with $N_t^{te}$ samples. The source and target domains are sampled from different marginal distributions, i.e., $P_s({x}) \neq P_t({x})$,  while the conditional distribution remains stable, i.e., $P_s(y|x) = P_t(y|x)$. The main goal of UDA is to minimize the distribution shift between  $P_s({x})$ and $P_t({x})$, assuming they share the same label space.

\subsection{General Approach}
The mainstream of UDA algorithms is to address the domain shift problem by finding domain invariant feature representation. Formally, given a feature extractor network $f_\theta: {X} \rightarrow {Z}$, which transforms the input space to the feature space, the UDA algorithm mainly optimizes the feature extractor network to minimize a domain alignment loss  $\mathcal{L}_{\mathrm{align}}$, aiming to mitigate the distribution shift between the source and target domains such that  $P_s(f_\theta(x)) = P_t(f_\theta(x))$. The domain alignment loss can either be estimated from a statistical distance measure or an adversarial discriminator network, which can be formalized as follows:

\begin{align}
    \mathcal{L}_{\mathrm{align}} = \min_{f_{\theta}, h_{\theta}} \ell (Z_s, Z_t),
\end{align}
where $\ell$ can be a statistical distance or an adversarial loss.

Concurrently, a classifier network $h_{\theta}$ is applied on top of the feature extractor network to map the encoded features to the corresponding class probabilities. Particularly, given the source domain features $Z_s$ generated from the feature extractor, we can calculate the output probabilities $\mathbf{p}_s = h_{\theta}(Z_s))$. Thus, the source classification loss can be formalized as follows
\begin{align}
    \mathcal{L}_{\mathrm{cls}}^{s}= -\mathbb{E}_{(\mathbf{x}_{s},y_{s}) \sim P_{s}} \sum_{k=1}^K  \mathbbm{1}_{[y_s = k]} \log \mathbf{p}_{s}^k , \label{eqn:src_cls}
\end{align}
where $\mathbbm{1}$ is the indicator function, which is set to be 1 when the condition is met and set to 0 otherwise.

Both the source classification loss  $\mathcal{L}_{\mathrm{cls}}^{s}$ and the domain alignment loss $\mathcal{L}_{\mathrm{align}}$ are jointly optimized to mitigate the domain shift while learning the source classification task, which can be expressed as

\begin{align}
    \min_{f_{\theta}, h_{\theta}} \mathcal{L}_{\mathrm{cls}}^{s} + \mathcal{L}_{\mathrm{align}}.
\end{align}

we refer to the composition of the the feature extractor $f_{\theta}$ and the classifier network $h_{\theta}$ as the model $m$, such that $m=h_\theta(f_\theta(\cdot))$.

\section{\adatime: A Benchmarking Approach for time series domain adaptation}
\label{sec:adatime}
\subsection{Framework Design}
The key motivation for our approach is to address the inconsistent evaluation schemes, datasets, and backbone networks. Such inconsistencies can boost the performance and be misattributed to the proposed UDA method. Therefore, we design our benchmarking framework to address these issues while ensuring fair evaluation across different UDA methods.  For example, to remove the effect of different backbone networks, we use the same backbone network when comparing different UDA methods. Additionally, we standardize the benchmarked datasets and their preprocessing schemes when evaluating any UDA method. Table \ref{t:summary} summarizes the existing experimental flaws and our corresponding design decision.

\begin{table}[ht]
    \centering
    \caption{Summary of the design choices in \adatime framework.}
    \resizebox{0.9\columnwidth}{!}{%
        \begin{tabular}{@{}cc@{}}
            \toprule
            Problem & Design Choice \\ \midrule
            Different datasets and cross-domain scenarios & Unified datasets and cross-domain scenarios for all the methods \\ 
            Inconsistent backbone networks & Same backbone network is fixed for all the methods \\
            Different training procedures & Unified the evaluation schemes and training procedure for all the methods \\
            \bottomrule
        \end{tabular}
    }
    \label{t:summary}
\end{table}

\subsection{Framework Overview}
In this work, we systematically evaluate different UDA algorithms on time series data, ensuring fair and realistic procedures. Fig.~\ref{fig:framework} shows the details of \adatime flow, which proceeds as follows. Given a dataset, we first apply our standard data preparation schemes on both domains, including slicing, splitting to train/test portions, and normalization. Subsequently, the backbone network extracts the source and target features $Z^{tr}_s$ and $Z^{tr}_t$ from the source training data $X^{tr}_s$ and target training data $X^{tr}_t$ respectively. The selected UDA algorithm is then applied to mitigate the distribution shift between the extracted features of the two domains. We generally categorize the adopted UDA algorithms into discrepancy- and adversarial-based approaches. Last, to set the hyper-parameters of the UDA algorithm, we consider three practical model selection approaches that do not require any target domain labels or allow for only few-shot labeled samples. These approaches are source risk (SRC), deep embedded evaluation risk (DEV) \cite{you2019towards}, and few-shot target risk (FST). Our evaluation pipeline standardizes experimental procedures, preventing extraneous factors from affecting performance, thus enabling fair comparison between different UDA methods.

The code of \adatime is publicly available for researchers to enable seamless evaluation of different UDA methods on time series data. Merging a new algorithm or dataset into \adatime will be just a matter of adding a few lines of code.

\subsection{Benchmarking Datasets}
We select the most commonly used time series datasets from two real-world applications, i.e., human activity recognition and sleep stage classification. The benchmark datasets span a range of different characteristics including complexity, type of sensors, sample size, class distribution, and severity of domain shift, enabling more broad evaluation. 

Table~\ref{tab:datasets} summarizes the details of each dataset, e.g., the number of domains, the number of sensor channels, the number of classes, the length of each sample, as well as the total number of samples in both training and test portions. The selected datasets are detailed as follows:

\subsubsection{UCIHAR}
UCIHAR dataset \cite{uciHAR_dataset} contains data from three sensors namely, accelerometer, gyroscope, and body sensors, that have been applied on 30 subjects. Each subject has performed six activities, i.e., walking, walking upstairs, downstairs, standing, sitting, and lying down.
Due to the aforementioned variability between subjects, we treat each subject as a separate domain. Here, we randomly selected five cross-domain scenarios out of a large number of cross-domain combinations, as in \cite{dskn,codats}. 

\subsubsection{WISDM}
In the WISDM dataset \cite{wisdm_dataset}, accelerometer sensors were applied to collect data from 36 subjects performing the same activities as in the UCIHAR dataset. However, this dataset can be more challenging because of the class imbalance issue in the data of different subjects. Specifically, data from some subjects may contain only samples from a subset of the overall classes (see Fig.~\ref{fig:class_distribution_har_wisdm} in the supplementary materials). Similar to the UCIHAR dataset, we consider each subject as a separate domain and randomly select five cross-domain scenarios.

\subsubsection{HHAR}
The Heterogeneity Human Activity Recognition (HHAR) dataset \cite{hhar_dataset} has been collected from 9 subjects using smartphone and smartwatch sensor readings. In our experiments, we use the same smartphone device, i.e., a Samsung smartphone, for all the selected subjects to reduce the heterogeneity. In addition, we consider each subject as one domain and form the five cross-domain scenarios from randomly selected subjects.

\subsubsection{SSC}
Sleep stage classification (SSC) problem aims to classify the electroencephalography (EEG) signals into five stages, i.e., Wake (W), Non-Rapid Eye Movement stages (N1, N2, N3), and Rapid Eye Movement (REM). We adopt Sleep-EDF dataset \cite{sleepEDF_dataset}, which contains EEG readings from 20 healthy subjects. Following previous studies, we select a single channel (i.e., Fpz-Cz) following previous studies \cite{attnSleep_paper} and include 10 subjects to construct the five cross-domain scenarios.

\subsubsection{MFD} The Machine Fault Diagnosis (MFD) dataset \cite{fd_dataset} has been collected by Paderborn University to identify various types of incipient faults using vibration signals. The data was collected under four different operating conditions, and in our experiments, each of these conditions was treated as a separate domain. We used five different cross-condition scenarios to evaluate the domain adaptation performance. Each sample in the dataset consists of a single univariate channel with 5120 data points following previous works~\cite{ragab2020adversarial,ccdg}.

\begin{table}[!htb]
    \centering
    \caption{Details of adopted datasets. Further details about selected cross-domain scenarios for each dataset can be found in the Appendix.}
    \begin{NiceTabular}{l|cccc|cc}
        \toprule
        Dataset & \# Users/Domains & \# Channels & \# Classes & Sequence Length & Training set & Testing set \\ \midrule
        UCIHAR & 30 & 9 & 6 & 128 & 2300 & 990 \\ 
        WISDM & 36 & 3 & 6 & 128 & 1350 & 720 \\
        HHAR & 9 & 3 & 6 & 128 & 12716 & 5218 \\ 
        SSC & 20 & 1 & 5 & 3000 & 14280 & 6130 \\ 
        MFD & 4 & 1 & 3 & 5120 & 7312 & 3604 \\ 
        \bottomrule
    \end{NiceTabular}
    \label{tab:datasets}
\end{table}

\subsection{Backbone Networks}
\label{sec:backnone}
In general, a UDA algorithm consists of a feature extractor network to extract the features from the input data, a classifier network to classify the features into different classes, and a domain alignment component to minimize the shift between domains. Here, we refer to the feature extractor as the backbone network. The backbone network transforms the data from the input space to the feature space, where UDA algorithms are usually applied. Thus, the backbone network can significantly influence the performance of the UDA method, independent of the actual domain adaptation component. Hence, standardizing the backbone network choice is necessary to compare different UDA methods fairly. However, some previous TS-UDA works adopted different backbone architectures when comparing against baseline methods, leading to inaccurate conclusions. 

To tackle this problem, we design \adatime to ensure the same backbone network is used when comparing different UDA algorithms, promoting fair evaluation protocols. Furthermore, to better select a suitable backbone network for TS-UDA application, we experiment with three different backbone architectures:
\begin{itemize}
    \item \textbf{1-dimensional convolutional neural network (1D-CNN)}: consists of three convolutional blocks, where each block consists of a 1D-Convolutional network, a BatchNorm layer, a non-linearity ReLU activation function, and finally, a MaxPooling layer \cite{eldele2021time,eldele2021adversarial}.
    
    \item \textbf{1-dimensional residual network (1D-ResNet)}: is a deep Residual Network that relies on a shortcut residual connection among successive convolutional layers~\cite{resnet_reference,fawaz2019deep}. In this work, we leveraged 1D-ResNet18 in our experiments.
    
    \item \textbf{Temporal convolutional neural network (TCN)}:  uses causal dilated convolutions to prevent information leakage across different convolutional layers and to learn temporal characteristics of time series data~\cite{bai2018empirical,thill2020time}.
\end{itemize}

These architectures are widely used for time series data analytics and differ in terms of their complexity and the number of trainable parameters.

\begin{table*}[ht]
    \caption{Summary of unsupervised domain adaptation algorithms implemented in \adatime.}
    \centering
    \begin{NiceTabular}{@{}l|ccccc@{}}
        \toprule
        \textbf{Algorithm} & \textbf{Application} & \textbf{Category} & \textbf{Distribution} & \textbf{Losses} &  \textbf{\textcolor{black}{Model Selection}} \\ \toprule 
        
        DDC & Visual & Discrepancy  & Marginal & MMD & \textcolor{black}{Target Risk}\\ \midrule

        Deep-Coral  &  Visual &Discrepancy  & Marginal &  CORAL & \textcolor{black}{Not Mentioned} \\ \midrule
        
        HoMM & Visual &Discrepancy  & Marginal & High-order MMD & \textcolor{black}{Not Mentioned}\\ \midrule
        
        MMDA  & Visual & Discrepancy  & Joint &  MMD,  & \textcolor{black}{Target Risk} \\ &&&& CORAL,\\ &&&& Entropy \\ \midrule
        
        DSAN & Visual & Discrepancy  & Joint & Local MMD & \textcolor{black}{Not Mentioned}\\ \midrule
        
        DANN & Visual & Adversarial  & Marginal & Domain Classifier, & \textcolor{black}{Source Risk} \\ 
        &&&& Gradient Reversal Layer \\ \midrule
        
        CDAN & Visual & Adversarial  & Joint & Conditional adversarial & \textcolor{black}{Importance  Weighting}\\ &&&& Domain Classifier \\ \midrule
        
        DIRT-T  &  Visual &Adversarial  & Joint &  Virtual adversarial & \textcolor{black}{Target Risk} \\ &&&& Entropy   \\ &&&& Domain Classifier\\ \midrule
        
        CoDATS & Time Series & Adversarial  & Marginal & Domain Classifier, & \textcolor{black}{Target Risk} \\
        &&&& Gradient Reversal Layer\\ \midrule
        
        AdvSKM  &  Time  Series &Adversarial  & Marginal &  Spectral Kernel & Target Risk \\ &&&& Adversarial MMD \\ \midrule
        SASA & Time Series & MMD & Marginal & Sparse Attention & Target Risk \\ &&&& MMD \\
        
         \bottomrule
    \end{NiceTabular}%

    \label{tbl:algorithms}
\end{table*}

\subsection{Domain Adaptation Algorithms}
While numerous UDA approaches have been proposed to address the domain shift problem \cite{da_review}, a comprehensive review of existing UDA methods is out of our scope. Besides including state-of-the-art methods proposed for time series data, we also included prevalent methods for visual UDA that can be adapted to time series. Overall, the implemented algorithms in \adatime can be broadly classified according to the domain adaptation strategy: discrepancy-based and adversarial-based methods. The discrepancy-based methods aim to minimize a statistical distance between source and target features to mitigate the domain shift problem \cite{ddc,deepcoral,HoMM}, while adversarial-based methods leverage a domain discriminator network that enforces the feature extractor to produce domain invariant features \cite{DANN,shu2018dirt}. Another way to classify UDA methods is based on what distribution is aligned distribution. Some algorithms only align the marginal distribution of the feature space \cite{mmd,deepcoral,DANN,codats,ddc,HoMM}, while others jointly align the marginal and conditional distributions \cite{CDAN,dsan,shu2018dirt,MMDA}, allowing fine-grained class alignment.

The selected UDA algorithms are as follows:
\begin{itemize}
    \item  \textbf{Deep Domain Confusion (DDC)} \cite{ddc}: minimizes the Maximum Mean Discrepancy (MMD) distance between the source and target domains.
    
    \item \textbf{Correlation Alignment via Deep Neural Networks (Deep CORAL)} \cite{sun2017correlation}: minimizes the domain shift by aligning second-order statistics of source and target distributions. 
    
    \item  \textbf{Higher-order Moment Matching (HoMM)} \cite{HoMM}: minimizes the discrepancy between different domains by matching higher-order moments of the source and target domains.
    
    \item\textbf{Minimum Discrepancy Estimation for Deep Domain Adaptation (MMDA)} \cite{MMDA}: combines both MMD and CORAL with conditional entropy minimization to address the domain shift.
    
    \item \textbf{Deep Subdomain Adaptation (DSAN)} \cite{dsan}: minimizes the discrepancy between source and target domains via a local maximum mean discrepancy (LMMD) that aligns relevant subdomain distributions. 
    
    \item \textbf{Domain-Adversarial Training of Neural Networks (DANN)} \cite{DANN}: leverages gradient reversal layer to adversarially train a domain classifier against feature extractor network. 

    \item \textbf{Conditional Adversarial Domain Adaptation (CDAN)} \cite{CDAN}: realizes a conditional adversarial alignment by incorporating the task knowledge with features during the domain alignment step.
    
     \item \textbf{A DIRT-T Approach to Unsupervised Domain Adaptation} \cite{shu2018dirt}: employs virtual adversarial training, conditional entropy, and teacher model to align the source and target domains.
      
    \item  \textbf{Convolutional deep Domain Adaptation model for Time Series data (CoDATS}) \cite{codats}: leverages adversarial training with weak supervision by a CNN network to improve the performance on time series data
    
    \item \textbf{ Adversarial Spectral Kernel Matching (AdvSKM)} \cite{dskn}: leverages adversarial spectral kernel matching to address the non-stationarity and non-monotonicity problem in time series data. 
    
    \item \textbf{Time Series Domain Adaptation via Sparse Associative Structure Alignment (SASA)} \cite{sasa}: exploits the causality property in time series data by aligning the sparse associative structure between the source and target domains.

\end{itemize}
Table~\ref{tbl:algorithms} summarizes the selected methods, showing the application for which each method was originally proposed, the classification of each method according to domain adaptation strategy (i.e., whether it relies on discrepancy measure or adversarial training), their classification based on the category of the aligned distribution (i.e., marginal or joint distribution), the losses in each method, and the risk that each UDA method adopted to tune its model hyper-parameters. It is worth noting that our \adatime mainly focuses on the time series classification problem, so we excluded methods proposed for time series prediction/forecasting.

\subsection{Model Selection Approaches}

\label{sec:risks_details}
Model selection and hyper-parameter tuning are long-standing non-trivial problems in UDA due to the absence of target domain labels. Throughout the literature, we find that the experimental setup in these works leverages target domain labels to select hyper-parameters, which violates the primary assumption of UDA. This is further clarified in Table \ref{tbl:algorithms}, where we find that five out of the 11 adopted UDA works use the target risk (i.e., target domain labels) in their experiments to select the hyper-parameters, while another three works use fixed hyper-parameters without describing how they are selected. To address this issue, we evaluate multiple realistic model selection approaches that do not require any target domain labels, such as source risk \cite{ganin2016domain} and Deep Embedded Validation (DEV) risk \cite{you2019towards}. In addition, we design a few-shot target risk that utilizes affordable few labeled samples from the target domain. In the following subsections explain the risk calculation for each model selection approach.

\subsubsection{Selection of Best Model}

As shown in Fig.~\ref{fig:framework}, given a set of $n$ candidate models $\mathcal{M}=(m_1, m_2, \dots, m_n)$ with different hyper-parameters. First, we calculate the corresponding risk value for each candidate model with respect to each model selection approach. Subsequently, we rank candidateThes based on their computed risk while selecting the model with minimum risk value. 
\begin{equation}
m_{best} = \min_{m \in \mathcal{M}} \mathcal{R}_{{*}}(m),
\end{equation}
where $m_{best}$ is the best model that achieves the minimum risk value, and $\mathcal{R}_{*}\in \{\mathcal{R}_{\text{SRC}}, \mathcal{R}_{\text{DEV}}, \mathcal{R}_{\text{FST}}, \mathcal{R}_{\text{TGT}} \}$ can be any of the model selection approaches described below.

\subsubsection{Source Risk (SRC)} 
In this approach, we select the candidate model that achieves the minimum cross-entropy loss on a test set from the source domain.
Therefore, this risk can be easily applied without any additional labeling effort as it relies on existing labels from the source domain  \cite{ganin2016domain}. Given the source domain test data $(x_s^{te}, y_s^{te})$, and a candidate model $m$, we calculate the corresponding source risk $\mathcal{R}_{SRC}$ as:

\begin{equation}
\mathcal{R}_{\text{SRC}}(m) = \E_{x_s \sim P_{s}(x)} \ell_{ce} (m(x_s^{te}),  y_s^{te}),
\end{equation}
where $\ell_{ce}$ is the cross-entropy loss. Despite the simplicity of the source risk, its effectiveness is mainly influenced by the sample size of source data and the severity of the distribution shift. When the distribution shift is large and the sample size of the source data is small, the source risk may be less effective than the target risk. However, the source risk can be estimated using only source labels, whereas the target risk requires labeled data from the target domain.

\subsubsection{DEV Risk} This approach  \cite{you2019towards} aims to find an unbiased estimator of the target risk. The key idea is to consider the correlation between the source and target features during the risk calculation. More specifically, the DEV method puts larger weights on the  source features highly correlated to the target features while giving lower weights to the less correlated features. To do so, an importance weighting scheme has been applied to the feature space. 
Given the source domain training features $Z_s^{tr}$, the source domain test set $Z_s^{te}$, and the target domain training features $Z_t^{tr}$, we first train a two-layer logistic regression model $r_\theta$ to discriminate between $Z_s^{tr}$ and $Z_t^{tr}$ (label features from $Z_s^{tr}$ as 1, and $Z_t^{tr}$ as 0), which can be formalized as follows

\begin{align}\label{d_loss}
\min_{r_\theta}\mathcal{L}_{d} =  \big[\log r_\theta(Z_s^{tr}))   +\log (1-r_\theta(Z_t^{tr})\big].
\end{align}
Subsequently, we leverage the trained $r_\theta$ to compute the importance weights $w_f$ for the source test set.
\begin{align}
w_f(x_s^{te})&=\frac{N_s^{tr}}{N_t^{tr}} \frac{1-r_\theta(Z_s^{te})}{r_\theta(Z_s^{te})},
\end{align}
where $\frac{N_s^{tr}}{N_t^{tr}}$ is sample size ratio of both domains. It is worth noting that the sample ratio parameter can be computed without any target labels. Given the importance weights for each test sample of the source domain, we compute the corresponding weighted cross-entropy loss, $L_w$, for the test samples of the source domain, which can be expressed as
\begin{align}
L_w&=\{w_f(x_s^{te}) \ell_{ce} (m(x_s^{te}), y_s^{te})\}_i^{N_s^{te}},
\end{align}
where $m$ is one candidate model. Given the weighted source loss $L_w$ and its corresponding importance weight $W=\left\{w_f\left(x_s^{te}\right)\right\}$, we compute the DEV risk as follows:
\begin{align}
\mathcal{R}_{\text{DEV}} &= \text{mean}(L_w) + \eta \text{mean}(W) - \eta,
\end{align}
where $\eta = - \frac{\text{Cov}(L_w)}{\text{Var}(W)}$ is the optimal coefficient. The DEV risk can be more effective than the source risk. However, we observed in our experiments that DEV may have unstable performance with smaller source and target datasets and adds additional computational overheads. Nevertheless, DEV risk is still a more practical solution than the target risk as it does not require any target labels.

\subsubsection{Target Risk (TGT)} This approach involves leaving out a large subset of target domain samples and their labels as a validation set and using them to select the best candidate model.
Using this risk naturally yields the best-performing hyper-parameters on the target domain. This can be seen as the upper bound for the performance of a UDA method. The target risk $\mathcal{R}_{\text{TGT}}$ is calculated as:
\begin{equation}
\mathcal{R}_{\text{TGT}} = \E_{x_t \sim P_{t}(x)} \ell_{ce} (m(x_t^{te}),  y_t^{te}).
\end{equation}
Even though this approach is impractical in unsupervised settings, it has been used for model selection in many previous UDA papers \cite{shu2018dirt,tri_training}.

\subsubsection{Few-Shot Target (FST) Risk} We propose the few-shot target risk as a more practical alternative to the target risk. Our goal in introducing the concept of few-shot target risk was to find a more practical and realistic model selection method for unsupervised domain adaptation. We reasoned that labeling a small number of samples, known as few-shot labeling, could be practical and affordable. Therefore, we used this approach to select the best model for the unsupervised domain adaptation problem. Formally speaking, we use a set of $q$ samples from the target domain as a validation set to select the best candidate model. The few-shot target risk $\mathcal{R}_{\text{FST}}$ is calculated as follows.

\begin{equation}
\mathcal{R}_{\text{FST}} = \frac{1}{q} \sum_{i=1}^{q} \ell_{ce} (m(x^{i}_t),  y^{i}_t).
\end{equation}

To our surprise, our experiments showed that the proposed few-shot risk was effective and performed comparably to the traditional target risk, despite being computed with only a few samples.

\subsection{Standardized Evaluation Scheme}

\subsubsection{Standardized Data Preprocessing} 
The preprocessing of time series data includes domain selection, data segmentation, data splitting, and data normalization. Specifically, the details are as follows:
\paragraph{Domain Selection}
For our closed-set domain adaptation experiments, we only consider datasets with similar source and target classes and representative samples for all categories. For example, in the WISDM dataset, some subjects (i.e., domains) may only perform certain activities during data collection, leading to the absence of some classes in the data from other subjects. Therefore, we exclude such subjects from our experiments and only consider subjects with complete data for all classes.
\paragraph{Data Segmentation}
During the data collection process, it is common for the sensor readings to be recorded in a single lengthy raw signal with a sequence length equal to the duration of the experiment. These long signals can present challenges for model training and can limit the number of available samples. To address these issues, we can use a sliding window technique to segment the data into shorter sequences. By doing so, we can both increase the number of samples and facilitate model training. For instance, in the domains of human activity recognition, sleep stage classification, and machine fault diagnosis, sliding windows of size 128, 3000, and 5120 are often used, respectively.
\paragraph{Data splitting}
Next, we divide the data into training and testing sets. Specifically, we split the data from each subject into stratified splits of 70\%/30\%, ensuring that the test set includes samples from all classes in the dataset. It is worth noting that we do not use a validation set for either the source or target domains. A validation set is typically used to select the best hyperparameters for the model, but we use four risk calculation methods that only require the source training data, source testing data, and target training data to select the best model.
\paragraph{Normalization}
Normalization is a crucial step in the training process of deep learning models, as it can help to accelerate convergence and improve performance. In this work, we apply Z-score normalization to both the training and testing splits of the data, using the following equation:

\begin{align}
x_i^{\text{normalize}}=\frac{x_i-x^{\text{mean}}}{x^{\text{std}}}, \quad i=1,2, \ldots, N
\label{eqn:normalization}
\end{align}
where $N=N_s$ for the source domain data and $N=N_t$ for the target domain data. Note that both the training and testing splits are normalized based on the training set statistics only~\cite{codats,dskn}.

\subsubsection{Standardized Training Scheme} 
All the training procedures have been standardized across all UDA algorithms. For instance, we train each model for 40 epochs, as performance tends to decrease with longer training. We report the model performance after the end of the last epoch. For model optimization, we use Adam optimizer with a fixed weight decay of 1e-4 and  $(\beta_1, \beta_2) = (0.5, 0.99)$. The learning rate is set to be a tunable hyperparameter for each method on each dataset. We exclude any learning rate scheduling schemes from our experiments to ensure that the contribution is mainly attributed to the UDA algorithm. 

\subsubsection{Hyper-parameter Search} 
\label{sec:sweep}
For each algorithm/dataset combination, we conduct an extensive random hyper-parameter search with 100 hyper-parameter combinations. The hyper-parameters are picked by a uniform sampling from a range of predefined values. Details about the specified ranges can be found in Table~\ref{tab:hyperparameter} of the supplementary materials. For each set of hyper-parameters, we calculate the values of the four risks for three different random seeds. We pick the model that achieves the minimum risk value for each model selection strategy. To calculate the FST risk, we use five samples per class from each dataset.

\subsubsection{Evaluation Metric}
Since time series data are usually imbalanced, some classes might be absent from some subjects (see Fig.~\ref{fig:class_distribution_har_wisdm} in the supplementary materials). The accuracy metric may also not be representative of the performance of the UDA methods. Therefore, we report macro F1-scores instead, considering how the data is distributed and avoiding predicting false negatives.

\section{Results and Discussions}
\label{sec:results}

In this section, we first investigate the contributions of different backbone networks to the performance of UDA algorithms. Subsequently, we study the performance of different model selection techniques on the benchmark datasets. Last, we discuss the main findings of our experiments. 
\begin{figure}[!htb]
  \centering
  \subfigure[CIHAR dataset]{\includegraphics[scale=0.33]{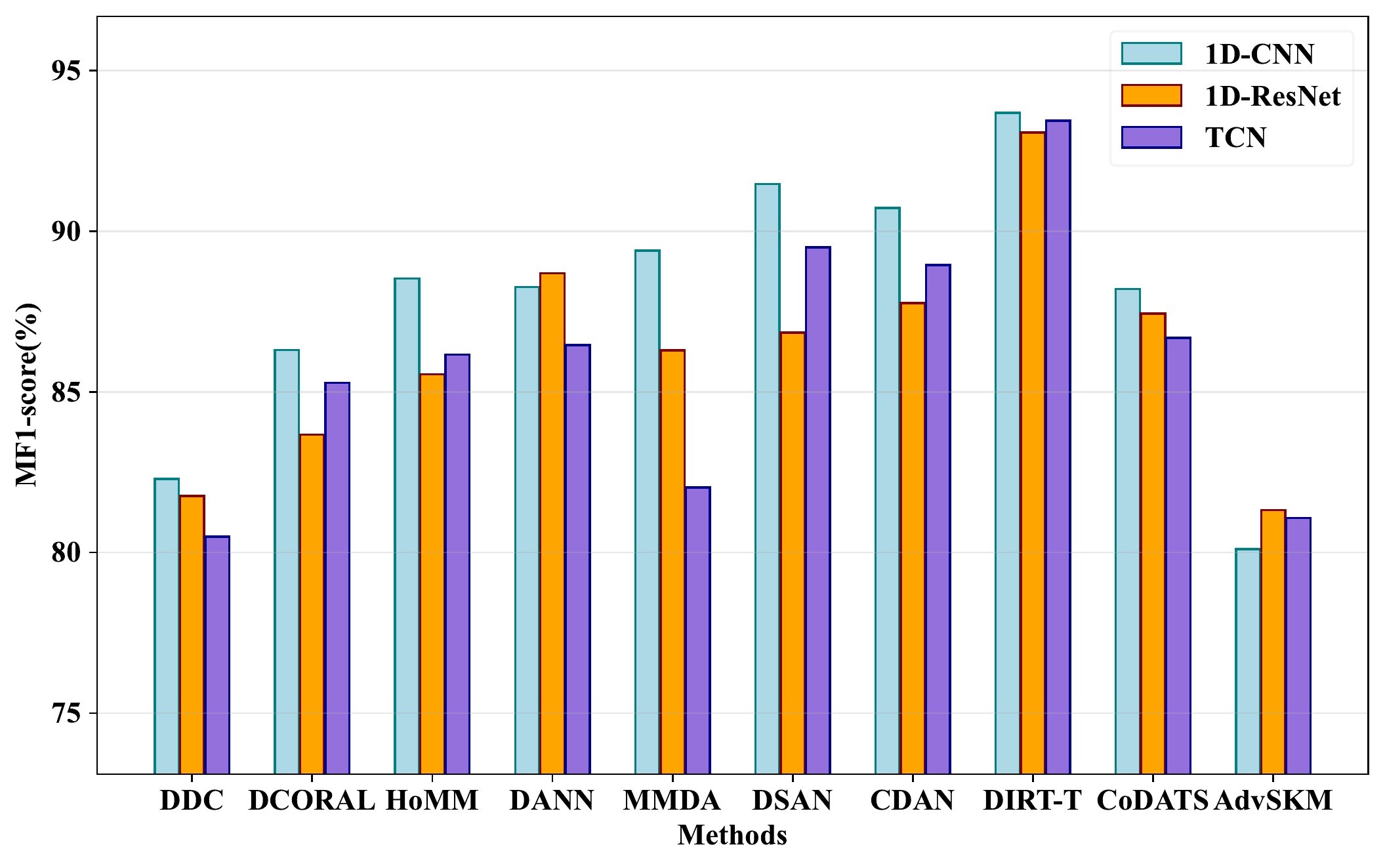} \label{subfig:comp_bkbone:har}}  \quad
  \subfigure[HHAR dataset]{\includegraphics[scale=0.33]{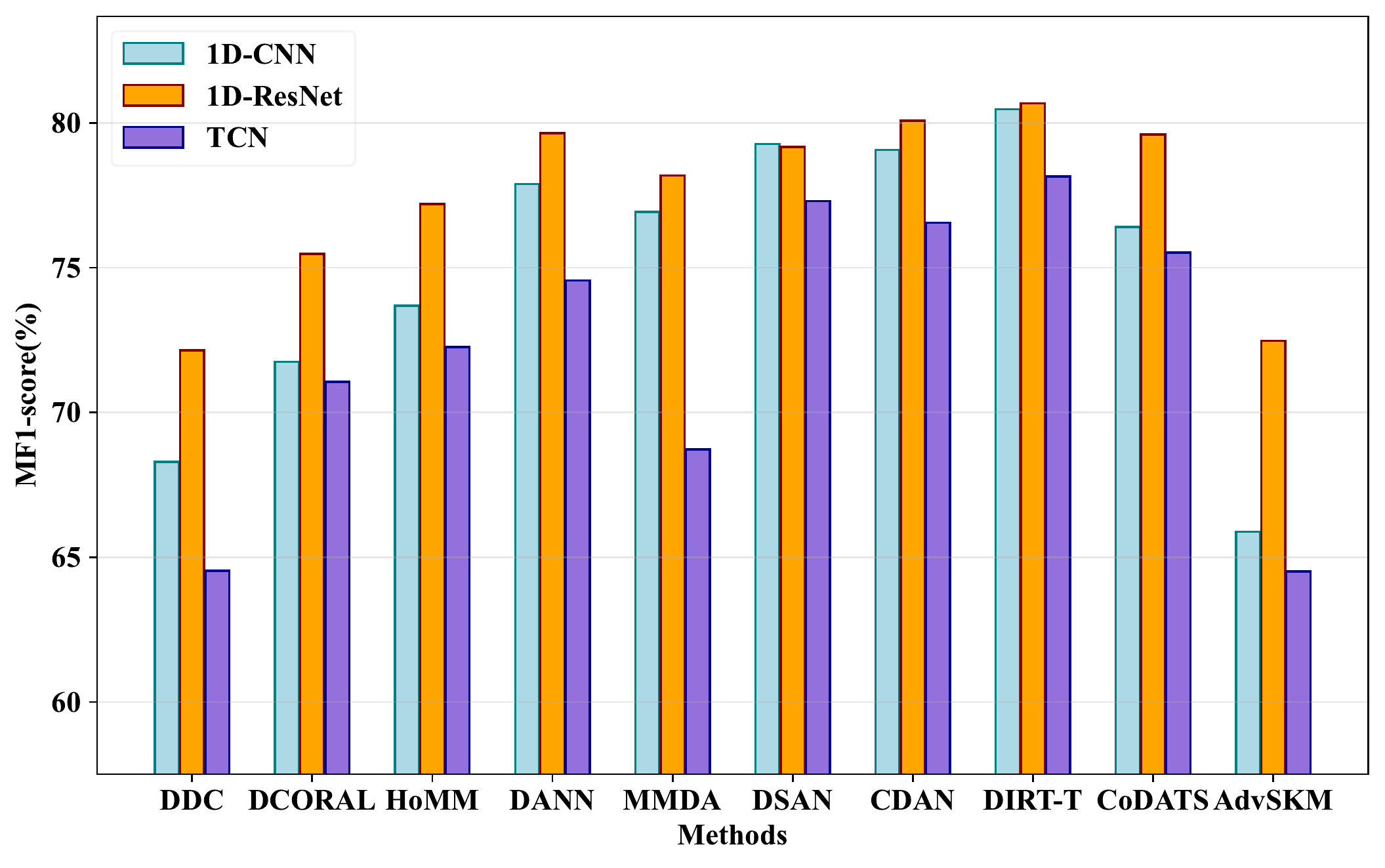} \label{subfig:comp_bkbone:hhar}}  \quad
  \caption{Comparison between 1D-CNN, 1D-ResNet, and TCN backbones applied on UCIHAR and HHAR datasets.}
  \label{fig:backbone}
\end{figure}

\subsection{Evaluation of Backbone Networks}
\label{sec:eval_backbone}
To investigate the impact of the backbone networks on the models' performance, we evaluate all the UDA algorithms under three different backbone networks. We employ 1D-CNN, 1D-ResNet, and TCN (described in Section~\ref{sec:backnone}) as backbone networks. To better evaluate the performance of different backbone networks, we experimented on datasets with different scales, i.e., the small-scale UCIHAR and the large-scale HHAR datasets. We reported the average performance of all cross-domain scenarios in the adopted datasets, as shown in Fig.~\ref{fig:backbone}. The absolute performance varies significantly across different backbone networks for the same UDA method.
Nonetheless, the relative performance between different UDA methods remains stable across the three backbone networks. For instance, while DIRT-T consistently performs best, DDC has the lowest performance with respect to all the other UDA methods, regardless of the utilized backbone networks. On the other hand, a comparison of the performance of 1D-CNN and 1D-ResNet on the UCIHAR and HHAR datasets reveals that 1D-CNN tends to perform better on the UCIHAR dataset. In contrast, 1D-ResNet tends to perform better on the HHAR dataset. This difference in performance may be due to the fact that a more complex model such as 1D-ResNet can better take advantage of the larger size of the HHAR dataset, leading to improved performance~\cite{fawaz2019deep}.

We also conducted additional experiments using Long-Short Term Memory (LSTM). We compared its performance to other CNN-based models on ten different methods for unsupervised domain adaptation on the UCIHAR dataset. Our results show that LSTM performed significantly worse than all the other CNN-based approaches. This may be due to the lower capacity of LSTM at modeling local patterns and producing class-discriminative features \cite{bai2018empirical}, as well as its difficulty in handling long-term dependencies~\cite{lea2016temporal}, which are common in many time series applications. Detailed results of the LSTM experiment are provided in the supplementary materials.

\subsection{Evaluation of Model Selection Strategies}
\label{sec:eval_model_selection}
In this experiment, we evaluate the performance of various model selection approaches, i.e., SRC, DEV, FST, and TGT (described in Section~\ref{sec:risks_details}) on the UDA methods. We first select the backbone network to be 1D-CNN due to its stable performance and computational efficiency. Then, for all the UDA algorithms, we choose the best model according to each model selection strategy while testing its performance on the target domain data. Table~\ref{tbl:main_results} shows the average F1-score performance across all the cross-domain scenarios spanning five different datasets (detailed versions can be found in Tables \ref{tbl:detailed_HAR}, \ref{tbl:detailed_wisd}, \ref{tbl:detailed_eeg}, \ref{tbl:detailed_HHAR}, and \ref{tbl:detailed_MFD} in the supplementary materials). The experimental results reveal the following conclusions. First, the performance of a UDA method can vary depending on the model selection strategy used. It is, therefore, essential to use the same model selection strategy when comparing different UDA methods. For example, in the WISDM dataset, the DANN method performs best when the model is selected based on the target risk, while the CoDATS method performs best when the model is selected based on the source risk. Second, in comparing various strategies for selecting a model with respect to the target risk, it appears that SRC and FST risks are effective for datasets with balanced class distributions, such as UCIHAR and HHAR. On the other hand, DEV risk tends to work better for imbalanced datasets, such as WISDM and MFD, although it may require more computational resources than other methods.

To summarize, when obtaining target labels is cost-prohibitive, both the source and DEV risks offer viable solutions as they do not require any target labels. The appropriate choice between the two depends on the dataset's characteristics and the computational resources' availability. While the DEV risk is more robust on class-imbalanced datasets, the source risk can be more computationally feasible. On the other hand, if only a small amount of labeled data from the target dataset is available, the few-shot risk can be the best choice as it achieves competitive performance with the target risk using only a small number of labeled samples, given that the target dataset has balanced classes.

\begin{table}[!ht]
    \centering
    \caption{The average results (for 10 cross-domain scenarios) according to the minimum risk value in terms of MF1-score applied on the 1D-CNN backbone.}
    \resizebox{\textwidth}{!}{
    \begin{NiceTabular}{l|c|ccccccccccc|c}
        \toprule
      Dataset & RISK & DDC & Deep & HoMM & DANN & MMDA & DSAN & CDAN & DIRT & CoDATS & AdvSKM & SASA & Avg/Risk \\ \midrule
        ~ & SRC & 81.56 & 85.52 & 86.89 & 87.31 & 84.72 & 87.91 & 85.98 & \textbf{89.64} & 84.30 & 78.10 & 82.81  & 85.19  \\ 
        \multirow{2}{*}{UCIHAR} & DEV & 74.52 & 79.08 & 78.69 & 84.26 & 88.49 & 88.70 & 84.10 & \textbf{92.34} & 78.69 & 80.01 & 81.68  & 82.89  \\ 
        ~ & FST & 81.64 & 86.30 & 88.52 & 86.59 & 87.96 & \textbf{91.46} & 88.87 & 90.79 & 86.99 & 80.00 & 84.02  & 86.91  \\ 
        ~ & TGT & 82.29 & 86.30 & 88.52 & 88.26 & 89.39 & 91.46 & 90.72 & \textbf{93.68} & 88.20 & 80.10 & 85.00 & 87.89  \\ \midrule
        
        ~ & SRC & 51.89 & 51.13 & 51.98 & 56.49 & 57.24 & 58.99 & 41.28 & 53.23 & \textbf{60.08} & 50.98 & 47.79  & 53.33  \\ 
        \multirow{2}{*}{WISDM} & DEV & 50.86 & 51.54 & 54.22 & 62.41 & 56.86 & \textbf{60.12} & 46.46 & 53.23 & 58.85 & 50.76 & 50.24 & 54.53  \\ 
        ~ & FST & 53.64 & 53.2 & 54.05 & 46.39 & 55.19 & \textbf{57.10} & 55.59 & 56.06 & 45.88 & 53.72 & 46.30  & 53.08  \\ 
        ~ & TGT & 53.78 & 54.19 & 56.92 & \textbf{62.41} & 59.82 & 61.08 & 55.59 & 59.59 & 62.06 & 54.82 & 53.34 & 58.03  \\ \midrule
        
        ~ & SRC & 68.14 & 69.99 & 70.00 & 76.11 & 66.94 & 75.67 & 77.32 & \textbf{78.56} & 73.89 & 65.32 & 74.17  & 72.44  \\ 
        \multirow{2}{*}{HHAR} & DEV & 65.07 & 65.16 & 67.68 & 61.74 & 69.32 & 76.67 & 76.91 & \textbf{80.34} & 72.6 & 65.20 & 72.20  & 70.07  \\ 
        ~ & FST & 67.91 & 71.75 & 73.69 & 73.56 & 75.10 & 77.98 & 76.06 & \textbf{79.90} & 71.28 & 64.61 & 75.73  & 73.18  \\ 
        ~ & TGT & 68.29 & 71.75 & 73.69 & 77.89 & 76.93 & 79.27 & 79.07 & \textbf{80.47} & 76.41 & 65.88 & 75.76 & 74.97  \\ \midrule
        
        ~ & SRC & 60.82 & 60.64 & 60.60 & 60.80 & 60.60 & 57.91 & 57.64 & 60.82 & 56.00 & \textbf{61.18} & 57.94  & 59.70  \\ 
        \multirow{2}{*}{SSC} & DEV & \textbf{60.84} & 56.16 & 54.52 & 60.80 & 58.63 & 59.37 & 57.99 & 52.49 & 54.61 & 56.19 & 57.94  & 57.16  \\ 
        ~ & FST & 60.86 & 60.84 & 60.61 & 60.80 & \textbf{61.14} & 58.60 & 53.95 & 59.40 & 55.64 & 61.10 & 56.37  & 59.29  \\ 
        ~ & TGT & 60.88 & 61.05 & 60.81 & 60.80 & \textbf{63.47} & 59.51 & 59.51 & 61.38 & 57.32 & 61.18 & 59.81 & 60.59  \\ \midrule
        
        ~ & SRC & 79.69 & 79.99 & 80.46 & 80.91 & 84.20 & 52.62 & 72.44 & \textbf{91.31} & 76.98 & 79.76 & 76.47 & 77.84  \\ 
        \multirow{2}{*}{MFD} & DEV & 80.21 & 78.58 & 79.78 & 81.95 & 82.45 & 80.11 & 78.34 & \textbf{87.37} & 80.29 & 81.19 & 77.59 & 81.03  \\ 
        ~ & FST & 79.69 & 79.99 & 80.17 & 80.91 & 84.20 & 74.82 & 64.89 & \textbf{91.31} & 81.73 & 79.76 & 77.74 & 79.75  \\ 
        ~ & TGT & 81.54 & 80.80 & 81.18 & 84.06 & 85.44 & 81.65 & 84.64 & \textbf{92.81} & 84.20 & 81.47 & 78.94 & 83.78 \\ 
        \bottomrule
    \end{NiceTabular}}
        \label{tbl:main_results}
\end{table}

\begin{table}[!htb]
\caption{Domain gap between source and target domains described by the difference between lower and upper-performance bounds on the 1D-CNN backbone.}
\label{tbl:domaingap}
\begin{NiceTabular}{@{}c|ccccc@{}}
\toprule
 & UCIHAR & WISDM & HHAR & SSC & MFD\\ \midrule
Same Domain (Target-only) & 100.00 & 98.02 & 98.55 & 72.09 & {99.39}\\
Cross-Domain (Source-only) & 65.94 & 48.60 & 63.07 & 51.67 & {72.51}\\
\midrule
Gap ($\delta$) & 37.32 & 49.44 & 33.86 & 18.38& 26.88\\ \bottomrule
\end{NiceTabular}
\end{table}

\begin{figure}[!hbtp]
  \centering
  \subfigure[UCIHAR dataset]{\includegraphics[scale=0.37]{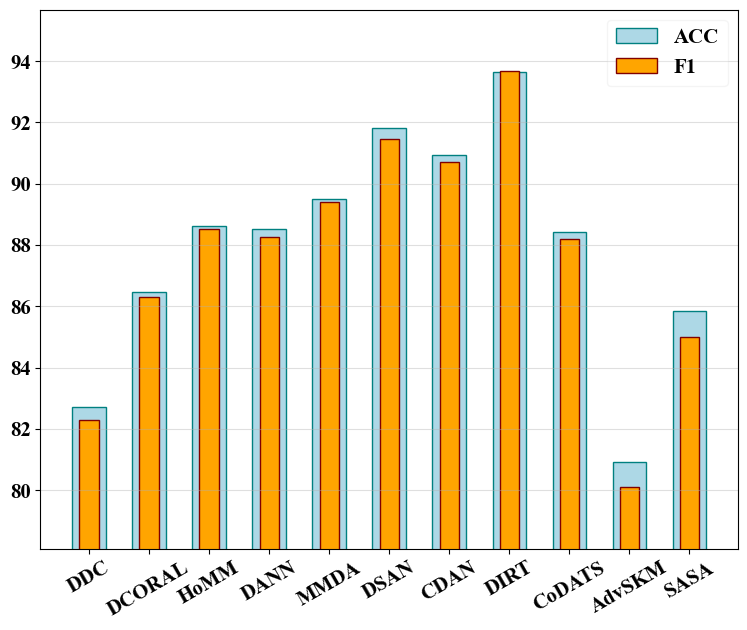} \label{subfig:comp:har}}  \quad
  \subfigure[WISDM dataset]{\includegraphics[scale=0.37]{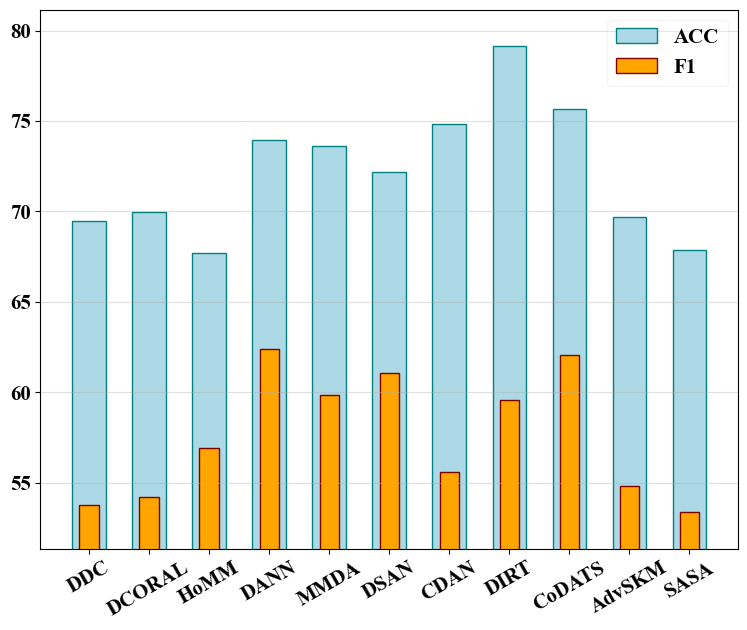} \label{subfig:comp:wisdm}}  \quad 
   \subfigure[HHAR dataset]{\includegraphics[scale=0.37]{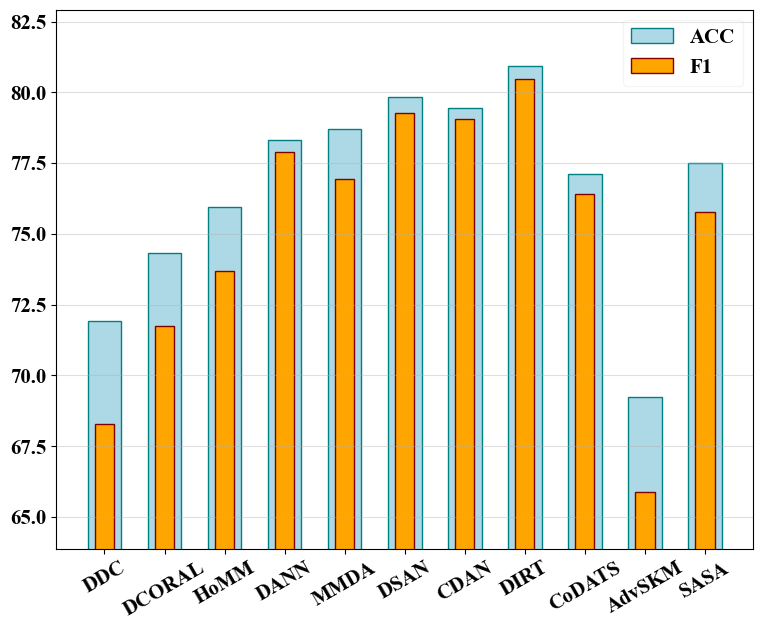} \label{subfig:comp:hhar}}  \quad
   \subfigure[SSC dataset]{\includegraphics[scale=0.37]{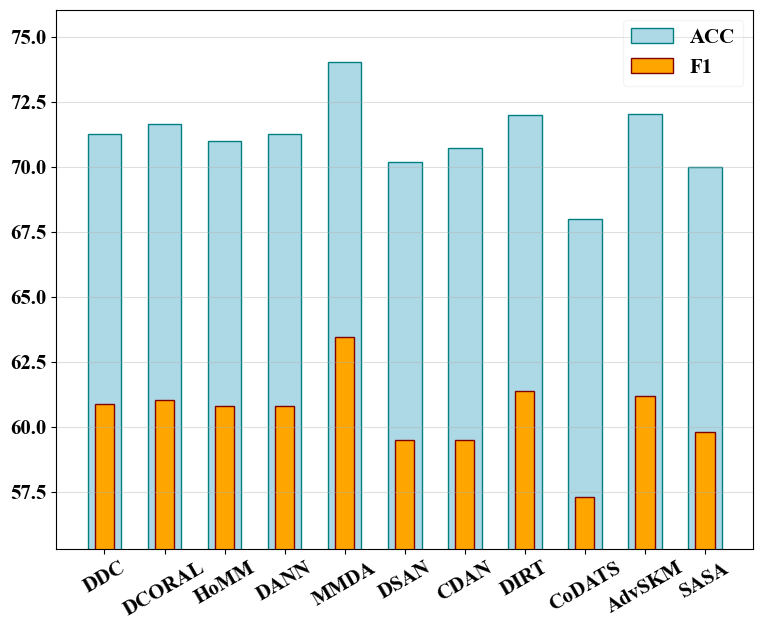} \label{subfig:comp:eeg}}  \quad
  \caption{Results of best models according to the target risk for different UDA methods in terms of accuracy and average macro F1-score.}
  \label{fig:acc_f1_comparison}
\end{figure}

\subsection{Discussions}
\label{sec:discussion}
\adatime provides a unified framework to evaluate different UDA methods on time series data. To explore the advantage of one UDA method over the others, we fixed the backbone network, the evaluation schemes, and the model selection strategy. We unveil the following insights. 

\textbf{Domain gap of different datasets.} We conducted the experiments on two small-scale datasets and two large-scale datasets. Regardless of the dataset size, all the adopted datasets suffer a considerable domain gap, as shown in Table~\ref{tbl:domaingap}. This table provides the results of the target-only experiment (i.e., training the model with target domain training set and testing it directly on the target domain test set) and the source-only experiment (i.e., training the model with source domain training set and testing it directly on the target domain test set). The backbone network for both experiments is the 1D-CNN network. While the source-only investigation represents the lower-bound performance, the target only represents the upper-bound performance, and their difference represents the domain gap.

\textbf{Visual UDA methods achieve comparable performance to TS-UDA methods on time series data.}
With further exploration of Table~\ref{tbl:main_results}, we find that, surprisingly, the performance of visual UDA methods is competitive or even better than TS-UDA methods. This finding is consistent for all the model selection strategies across the benchmarking datasets. For example, with the TGT risk value, we find that the methods proposed for visual applications such as DIRT-T and DSAN perform better than CoDATS and AdvSKM on the four datasets. A possible explanation is that all the selected UDA algorithms are applied on the vectorized feature space generated by the backbone network, which is independent of the input data modality. This finding suggests that visual UDA algorithms can be strong baselines for TS-UDA with a standard backbone network.

\textbf{Methods with joint distribution alignment tend to perform consistently better.}
Table \ref{tbl:algorithms} illustrates that some methods address the marginal distribution, such as MMD, CORAL, and HoMM, while the others handle the joint distributions (i.e., both marginal and conditional distributions concurrently) such as DIRT-T, MMDA, and DSAN. The results shown in Table~\ref{tbl:main_results} suggest that the methods addressing the joint distribution outperform those handling the marginal distribution. For example, the best-performing method, as selected by the TGT risk, is DIRT-T in UCIHAR and HHAR datasets and MMDA in the case of the SSC dataset. 
Similarly, with respect to different risks,  DIRT-T, MMDA, and DSAN interchangeably achieve the best performance across the benchmarking datasets. Hence, considering the conditional distribution when designing the UDA algorithm benefits the performance.

\textbf{Accuracy metric should not be used to measure performance for imbalanced data.} 
It is well known that accuracy is not a reliable metric for evaluating the performance of classifiers on imbalanced datasets. Despite this, many existing TS-UDA methods use accuracy to measure performance in their evaluations\cite{har_sys,codats,dskn}. Our experiments reveal that only using accuracy or F1-score alone can lead to inconsistent results on imbalanced datasets such as WISDM. This highlights the need to consider the imbalanced nature of most time series data when evaluating classifier performance. To illustrate this point, we present the results of our experiments in terms of both accuracy and F1-score on four datasets: WISDM, SSC, UCIHAR, and HHAR. WISDM and SSC are imbalanced, while UCIHAR and HHAR are mostly balanced.
As shown in Fig.~\ref{fig:acc_f1_comparison}, on the imbalanced WISDM dataset (Fig.~\ref{subfig:comp:wisdm}), CDAN achieves higher accuracy than some other methods such as DDC, MMDA, and DSAN, but has one of the worst performance in terms of F1-score. In contrast, the results on the balanced UCIHAR dataset (Fig.~\ref{subfig:comp:har}) show that accuracy can still be a representative performance measure and is similar to F1-score.
Therefore, we recommend using the F1-score as a performance measure in all TS-UDA experiments.

\textbf{Effect of labeling budget on the Few-shot risk performance}
To investigate the influence of the number of labeled samples on few-shot learning performance, we conducted additional experiments using 10 and 15 labeled samples per class on the UCIHAR dataset. The results in Table \ref{tbl:few_shots} show that the overall performance was relatively consistent regardless of the size of the few-shot labeled sample, indicating the robustness of few-shot learning to this hyperparameter.

\begin{table}[hb]
\caption{Model selection based on few-shot target risk  under a different budget of labeled samples on the UCIHAR dataset.}
\resizebox{\textwidth}{!}{
\begin{NiceTabular}{@{}l|cccccccccccc@{}}
\toprule
{{\# Labels}} & {DDC} & {DCORAL} & {HoMM} & {DANN} & {MMDA} & {DSAN} & {CDAN} & {DIRT-T} & {CoDATS} & {AdvSKM} & {SASA} & {Avg/risk} \\ \midrule
5 SHOTS & 81.64 & 86.3 & 88.52 & 86.59 & 87.96 & 91.46 & 88.87 & 90.79 & 86.99 & 80.00 &84.02 &86.65 \\
10 SHOTS & 81.53 & 86.3 & 88.52 & 87.25 & 88.69 & 90.97 & 88.87 & 90.79 & 87.73 & 80.00 & 84.02&86.79 \\
15 SHOTS & 81.53 & 86.3 & 88.52 & 86.13 & 87.96 & 91.46 & 88.87 & 90.79 & 87.73 & 80.00 & 84.02&86.66 \\ \bottomrule
\end{NiceTabular}}
\label{tbl:few_shots}
\end{table}

\textbf{Limitations and future works}
One limitation of our study is that it only focuses on time series classification. In future work, we plan to extend the scope of our benchmarking to include time series regression and forecasting tasks. Additionally, we have only considered the closed-set domain adaptation scenario, where the source and target classes are similar. In future work, we aim to consider partial and open-set domain adaptation scenarios, which are common in time series applications and involve varying classes between the source and target domains.

\section{Conclusions and Recommendations}
\label{sec:conclusion}
In this work, we provided \adatime, a systematic evaluation suite for evaluating existing domain adaptation methods on time series data. To ensure fair and realistic evaluation, we standardized the benchmarking datasets, evaluation schemes, and backbone networks among domain adaptation methods. Moreover, we explored more realistic model selection approaches that work without target domain labels or with only a few-shot labeled samples. Based on our systematic study, we provide some recommendations as follows. First, visual UDA methods can be applied to time series data and are strong candidate baselines. Second, we can rely on more realistic model selection strategies that do not require target domain labels, such as source risk and DEV risk, to achieve reliable performance.
Third, we recommend conducting experiments with large-scale datasets to obtain reliable results by fixing the backbone network among different UDA baselines. We also suggest adopting the F1-score instead of accuracy as a performance measure to avoid any misleading results with imbalanced datasets. Lastly, we believe that incorporating time series-specific domain knowledge into the design of UDA methods has the potential to be beneficial moving forward.

\section{Acknowledgments}
This work was supported by the Agency of Science Technology and Research under its AME Programmatic (Grant No. A20H6b0151) and its Career Development Award (Grant No. C210112046). The work was also supported by the MOE Academic Research (Grant No: MOE2019-T2-2-175). 

\bibliographystyle{unsrt}
\bibliography{ref}

\newpage

\renewcommand\thefigure{S\arabic{figure}}
\renewcommand{\thesection}{S\Roman{section}} 
\renewcommand\thetable{S\arabic{table}}
\setcounter{figure}{0}
\setcounter{table}{0}
\setcounter{section}{0}

\section{Class distribution of different subjects}
\begin{figure}[htp]
  \centering
  \subfigure[UCIHAR dataset]{\includegraphics[scale=0.14]{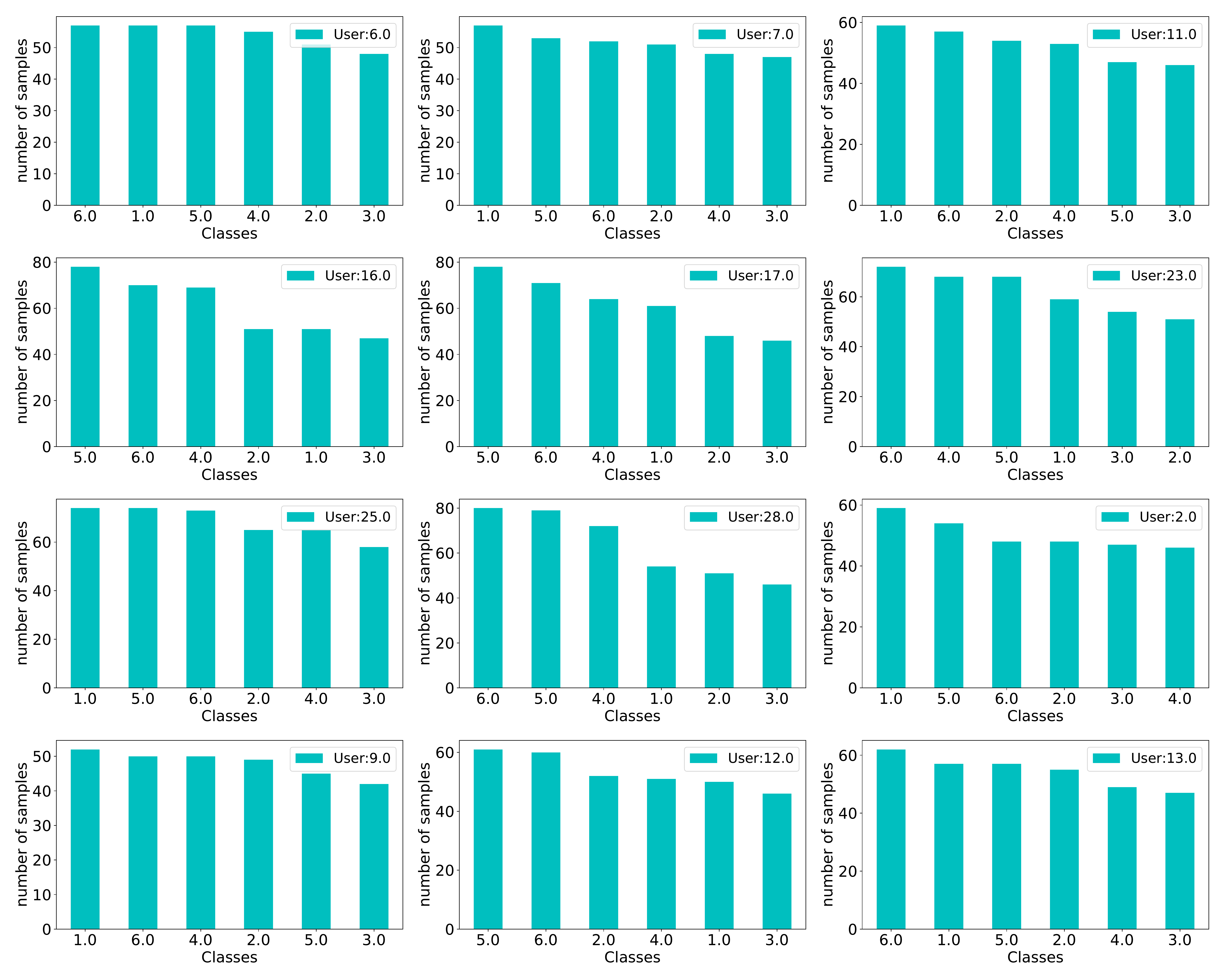} \label{fig:class_distribution_har_wisdm:har}}\quad
  \subfigure[WISDM dataset]{\includegraphics[scale=0.14]{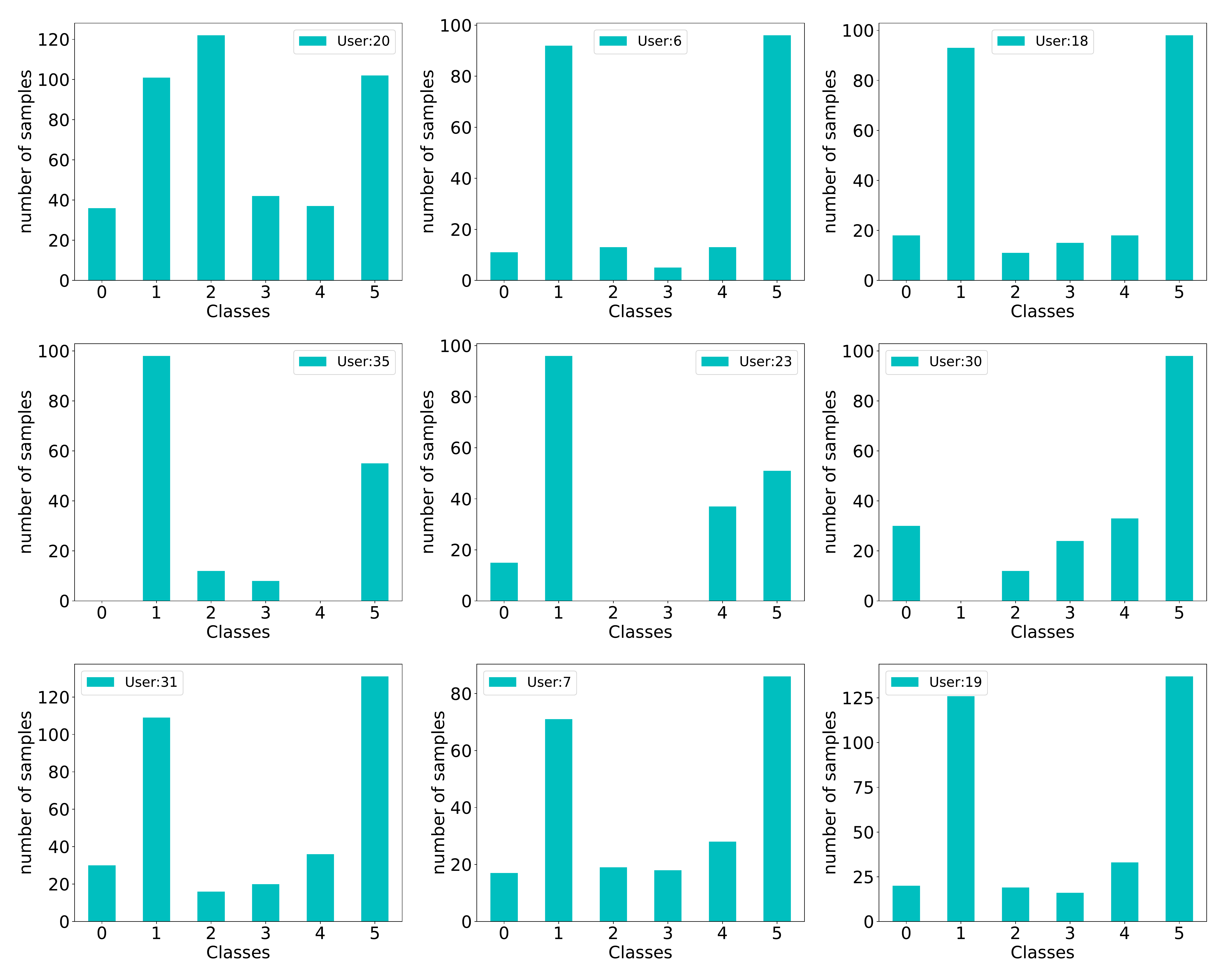} \label{fig:class_distribution_har_wisdm:wisdm}}
  \caption{Class distribution of selected subjects among different datasets}
  \label{fig:class_distribution_har_wisdm}
\end{figure}

This section visualizes the class distribution for each selected subject across all datasets. Specifically, Figure S\ref{fig:class_distribution_har_wisdm:har} depicts the class distribution of subjects in the UCIHAR dataset, where it is noted that all subjects possess data for every class. Conversely, as shown in Figure S\ref{fig:class_distribution_har_wisdm:wisdm}, certain subjects within the WISDM dataset lack data for select classes.

\begin{table}[h]
\centering \caption{Details of hyper-parameter tuning setup. }

{\setlength\doublerulesep{0.4pt}   

\begin{tabular}{*{3}{l}}
\toprule[1pt]\midrule[0.3pt]
\textbf{Method} & \textbf{Hyperparameter}   & \textbf{Range}   \\ 
\midrule

& {\parbox{4cm}{Learning Rate }}
& {\parbox{2cm}{$10^{-3}$ to $10^{0}$}} \\ \midrule
DDC
& \multirow{2}{*}{\parbox{4cm}{ MMD loss  \\ Classification loss }}
& \multirow{2}{*}{\parbox{2cm}{$10^{-2}$ to $10^{1}$\\ $10^{-1}$ to $10^{1}$}} \\
&\\ \midrule
Deep CORAL
& \multirow{2}{*}{\parbox{4cm}{ Coral loss  \\ Classification loss }}
& \multirow{2}{*}{\parbox{2cm}{$10^{-2}$ to $10^{1}$\\ $10^{-1}$ to $10^{1}$}} \\
&\\
\midrule

HoMM 
& \multirow{2}{*}{\parbox{4cm}{ High-order-MMD loss  \\ Classification loss }}
& \multirow{2}{*}{\parbox{2cm}{$10^{-2}$ to $10^{1}$\\ $10^{-1}$ to $10^{1}$}} \\
&\\
\midrule
MMDA 
& \multirow{2}{*}{\parbox{4cm}{ MMD loss  \\Coral Loss \\ Conditional loss \\ Classification loss}}
& \multirow{2}{*}{\parbox{2cm}{$10^{-2}$ to $10^{1}$\\ $10^{-2}$ to $10^{1}$ \\ $10^{-2}$ to $10^{1}$ \\ $10^{-1}$ to $10^{1}$}} \\\\\\\\
\midrule

DSAN 
& \multirow{2}{*}{\parbox{4cm}{ Local MMD loss  \\ Classification loss }}
& \multirow{2}{*}{\parbox{2cm}{$10^{-2}$ to $10^{1}$\\ $10^{-2}$ to $10^{1}$}} \\
&\\
\midrule
DANN 
& \multirow{2}{*}{\parbox{4cm}{ MMD loss  \\ Classification loss }}
& \multirow{2}{*}{\parbox{2cm}{$10^{-2}$ to $10^{1}$\\ $10^{-1}$ to $10^{1}$}} \\
&\\
\midrule

CDAN
& \multirow{2}{*}{\parbox{4cm}{Adversarial loss  \\ Conditional loss  \\ classification loss   }}
& \multirow{2}{*}{\parbox{2cm}{ $10^{-2}$ to $10^{1}$ \\ $10^{-2}$ to $10^{1}$ \\
$10^{-1}$ to $10^{1}$}} \\\\\\ \midrule

DIRT-T
& \multirow{2}{*}{\parbox{4cm}{Adversarial loss  \\ Conditional loss  \\ virtual adversarial  \\ Discriminator steps \\ classification loss  }}
& \multirow{2}{*}{\parbox{2cm}{ $10^{-2}$ to $10^{1}$ \\ $10^{-2}$ to $10^{1}$ \\ \ $10^{-2}$ to $10^{1}$ \\ $10^{-2}$ to $10^{1}$ \\ $10^{-1}$ to $10^{1}$}} \\
\\\\\\\\ \midrule

CODATS 
& \multirow{2}{*}{\parbox{4cm}{Adversarial loss  \\ classification lossw}}
& \multirow{2}{*}{\parbox{2cm}{$10^{-2}$ to $10^{1}$\\ $10^{-1}$ to $10^{1}$}} \\\\ \midrule

AdvSKM 
& \multirow{2}{*}{\parbox{4cm}{Adversarial MMD loss \\ Classification loss  }}
& \multirow{2}{*}{\parbox{2cm}{$10^{-2}$ to $10^{1}$\\ $10^{-1}$ to $10^{1}$}} \\\\ \midrule

SASA 
& \multirow{2}{*}{\parbox{4cm}{Domain loss \\ Classification loss  }}
& \multirow{2}{*}{\parbox{2cm}{$10^{-2}$ to $10^{1}$\\ $10^{-1}$ to $10^{1}$}} \\\\

\midrule[0.3pt]
\bottomrule[1pt]
\end{tabular}
}
\label{tab:hyperparameter}
\vspace{-1em}
\end{table}

\section{Detailed parameter ranges for the hyper-parameter search}
We provide the detailed ranges for each parameter among all selected domain adaptation methods, which are shown in Table~\ref{tab:hyperparameter}. 
We tuned the learning rate from the same range for all the UDA algorithms while we chose different ranges for each specific loss in its prospective UDA method.

\section{Backbone Network Architecture}
\begin{figure}[h]
    \centering
    \includegraphics{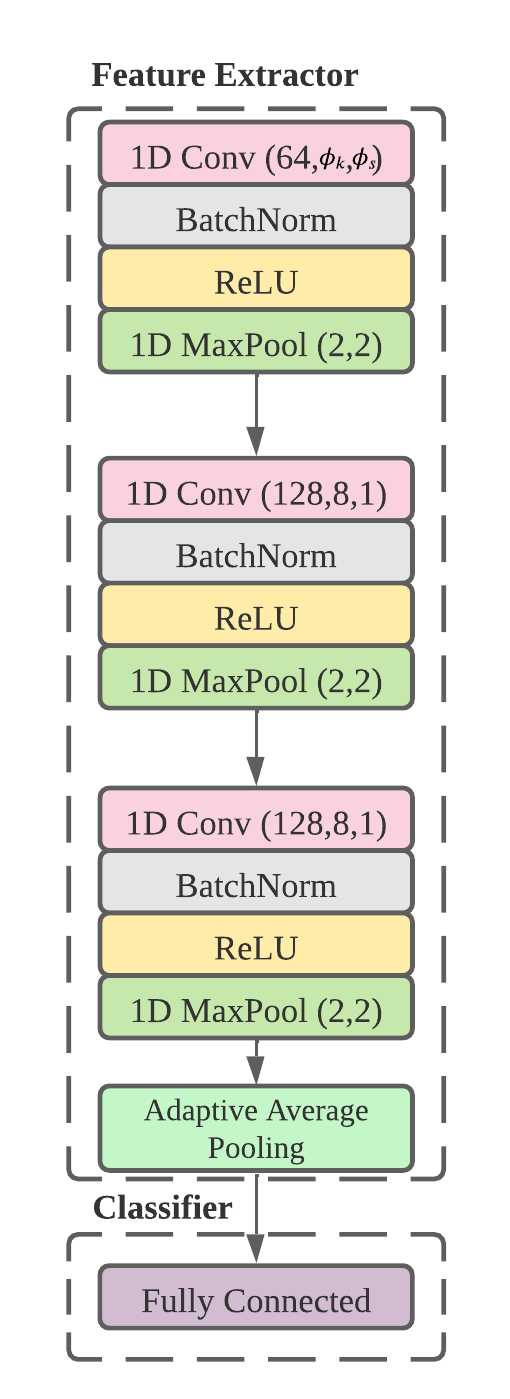}
    \caption{Backbone network of 1D-CNN, where $\phi_k$ is the kernel size and $\phi_s$ is the stride.}
    \label{fig:detailed_backbone}
\end{figure}

Fig.~\ref{fig:detailed_backbone} describes the detailed structure of the 1D-CNN network that we used as a backbone in our \adatime. It consists of a 3-block CNN, and each block has a 1D convolutional layer, followed by a 1D batch norm layer, a ReLU function for non-linearity, and finally, a 1D MaxPooling layer. The first convolutional layer in the first block has a kernel size of $\phi_k$ and a stride $\phi_s$, and those differ according to the dataset. The details of their values for each dataset can be found on the Github repository.
Regarding the 1D-ResNet-18, we deployed the one mentioned in \cite{resnet_reference}, which is now a standard architecture.

\begin{table}[!htb]
    \caption{Average F1-score of different backbone networks on UCIHAR dataset over 10 cross-domain scenarios}
    \begin{NiceTabular}{@{}l|cccccccccc@{}}
        \toprule
        Backbone & {DDC} & {DCORAL} & {HoMM} & {DANN} & {MMDA} & {DSAN} & {CDAN} & {DIRT-T} & {CoDATS} & {AdvSKM} \\ \midrule
        LSTM & 51.69 & 52.08 & 53.89 & 60.92 & 52.03 & 49.07 & 56.66 & 66.16 & 58.07 & 52.80 \\ \midrule
        1D-CNN & \textbf{82.29} & \textbf{86.30} & \textbf{88.52} & 88.26 & \textbf{89.39} & \textbf{91.46} & \textbf{90.72} & \textbf{93.68} & \textbf{88.20} & 80.10 \\
        1D-RESNET18 & 81.76 & 83.67 & 85.54 & \textbf{88.69} & 86.29 & 86.84 & 87.76 & 93.07 & 87.43 & \textbf{81.32} \\
        TCN & 80.49 & 85.28 & 86.16 & 86.45 & 82.02 & 89.49 & 88.94 & 93.43 & 86.68 & 81.07 \\ \bottomrule
    \end{NiceTabular}
    \label{tbl:lstm_backbone}
\end{table}

\section{Performance of LSTM-based backbone network}
Recurrent Neural Networks (RNNs) are widely used for time series forecasting and regression due to their ability to learn the temporal dynamics of time series signals. However, there are several limitations to using RNNs for time series classification tasks. First, compared to CNN-based approaches, RNNs are less effective at modeling local patterns and producing class-discriminative features, which can negatively impact their classification performance. Second, RNNs struggle to handle long-term dependencies common in many time series applications. For example, the sleep stage classification (SSC) dataset has a sequence length of 3000 timesteps per sample, which can be challenging for RNNs to process. Lastly, while CNN-based backbones can be efficiently trained using parallel computations, RNNs require sequential computation, leading to longer training times. Given these limitations, we decided to focus on CNN-based backbones in our experiments. However, we also conducted experiments using an LSTM backbone network. We compared its performance to other CNN-based backbones on ten different UDA methods on the UCIHAR dataset, as shown in Table~\ref{tbl:lstm_backbone}. The results of our experiments show that the LSTM backbone performs significantly worse than all the other CNN-based backbones, demonstrating the deficiency of RNNs on time series classification tasks.

\section{Impact of model selection on UDA performance}

\begin{table}[!ht]
\caption{Ranking of different UDA methods with respect to different model selection strategies on SSC dataset}
\label{tab:model_select}
\begin{NiceTabular}{@{}c|ccccccccccc@{}}
\toprule
\multicolumn{1}{l|}{Risks} & DDC & DCORAL & HoMM & DANN & MMDA & DSAN & CDAN & DIRT & CoDATS & AdvSKM & SASA\\ \midrule
SRC & 2 & 5 & 6 & 4 & 6 & 9 & 10 & 2 & 11 & 1 & 8 \\
DEV & 1 & 8 & 10 & 2 & 4 & 3 & 5 & 11 & 9 & 7 & 6 \\
FST & 3 & 4 & 6 & 5 & 1 & 8 & 11 & 7 & 10 & 2 & 9 \\
TGT & 5 & 4 & 6 & 7 & 1 & 9 & 9 & 2 & 11 & 3 & 8 \\ \bottomrule
\end{NiceTabular}
\end{table}

To demonstrate the impact of model selection on performance, we evaluated various unsupervised domain adaptation (UDA) methods on the SSC dataset using multiple model selection strategies. The results in Table~\ref{tab:model_select} reveal that the ranking of the UDA methods varies according to the model selection strategy employed. This demonstrates the importance of carefully considering the appropriate model selection strategy for the domain adaptation task.

\section{Detailed results of our Framework}
This subsection provides detailed results of 10 scenarios for all the datasets. In specific, Tables \ref{tbl:detailed_HAR}, \ref{tbl:detailed_wisd}, \ref{tbl:detailed_eeg}, \ref{tbl:detailed_HHAR}, and \ref{tbl:detailed_MFD} show the mean and the standard deviation for each cross-domain scenario in UCIHAR, WISDM, SSC, HHAR, and MFD dataset respectively. 

\begin{table}[!ht]
    \centering
    \caption{UCIHAR Dataset}
    \label{tbl:detailed_HAR}
    \resizebox{\textwidth}{!}{
    \begin{NiceTabular}{l|c|c|c|c|c|c|c|c|c|c|c}
    \toprule
        Algorithm &RISK & 2$\rightarrow$11 & 6$\rightarrow$23 & 7$\rightarrow$13 & 9$\rightarrow$18 & 12$\rightarrow$16 & 18$\rightarrow$27 & 20$\rightarrow$5 & 24$\rightarrow$8 & 28$\rightarrow$27 & 30$\rightarrow$20  \\ \midrule
        ~ & SRC & 58.23 $\pm$ 2.82 & 97.7 $\pm$ 1.15 & 71.48 $\pm$ 6.51 & 92.39 $\pm$ 1.84 & 97.57 $\pm$ 3.36 & 100.0 $\pm$ 0.0 & 82.44 $\pm$ 1.46 & 82.05 $\pm$ 3.23 & 85.97 $\pm$ 3.05 & 47.8 $\pm$ 6.18   \\ 
        \multirow{2}{*}{DDC} & DEV & 59.4 $\pm$ 2.21 & 83.11 $\pm$ 11.36 & 66.15 $\pm$ 5.3 & 94.83 $\pm$ 3.39 & 82.49 $\pm$ 7.46 & 77.95 $\pm$ 19.18 & 68.73 $\pm$ 15.35 & 84.75 $\pm$ 4.44 & 76.88 $\pm$ 7.84 & 50.95 $\pm$ 3.81 \\ 
        ~ & FST & 58.23 $\pm$ 2.82 & 97.7 $\pm$ 1.15 & 71.78 $\pm$ 6.01 & 92.39 $\pm$ 1.84 & 97.57 $\pm$ 3.36 & 100.0 $\pm$ 0.0 & 82.44 $\pm$ 1.46 & 82.05 $\pm$ 3.23 & 85.97 $\pm$ 3.05 & 48.23 $\pm$ 5.03 \\ 
        ~ & TGT & 62.13 $\pm$ 9.26 & 97.35 $\pm$ 1.7 & 71.04 $\pm$ 6.97 & 92.86 $\pm$ 2.21 & 97.37 $\pm$ 2.33 & 99.34 $\pm$ 1.15 & 82.93 $\pm$ 1.05 & 86.49 $\pm$ 1.02 & 86.94 $\pm$ 4.62 & 46.4 $\pm$ 1.92  \\ \midrule

        ~ & SRC & 64.63 $\pm$ 6.44 & 98.08 $\pm$ 0.66 & 73.68 $\pm$ 7.31 & 95.59 $\pm$ 4.32 & 97.89 $\pm$ 2.3 & 100.0 $\pm$ 0.0 & 80.59 $\pm$ 2.58 & 91.14 $\pm$ 0.56 & 87.27 $\pm$ 4.31 & 66.32 $\pm$ 2.4 \\ 
        \multirow{2}{*}{DCORAL} & DEV & 57.98 $\pm$ 2.56 & 78.1 $\pm$ 11.66 & 69.31 $\pm$ 0.92 & 93.88 $\pm$ 1.92 & 87.28 $\pm$ 7.48 & 88.88 $\pm$ 9.71 & 81.78 $\pm$ 1.75 & 88.38 $\pm$ 5.18 & 89.36 $\pm$ 2.86 & 55.87 $\pm$ 5.81 \\ 
        ~ & FST & 62.66 $\pm$ 3.02 & 98.47 $\pm$ 1.33 & 80.91 $\pm$ 8.62 & 94.18 $\pm$ 3.29 & 99.66 $\pm$ 0.59 & 100.0 $\pm$ 0.0 & 78.8 $\pm$ 1.77 & 92.61 $\pm$ 2.89 & 85.22 $\pm$ 4.02 & 70.47 $\pm$ 5.16  \\ 
        ~ & TGT & 62.66 $\pm$ 3.02 & 98.47 $\pm$ 1.33 & 80.91 $\pm$ 8.62 & 94.18 $\pm$ 3.29 & 99.66 $\pm$ 0.59 & 100.0 $\pm$ 0.0 & 78.8 $\pm$ 1.77 & 92.61 $\pm$ 2.89 & 85.22 $\pm$ 4.02 & 70.47 $\pm$ 5.16   \\ \midrule

        ~ & SRC & 65.58 $\pm$ 6.42 & 99.62 $\pm$ 0.67 & 76.97 $\pm$ 8.83 & 97.12 $\pm$ 4.98 & 96.61 $\pm$ 3.0 & 100.0 $\pm$ 0.0 & 81.79 $\pm$ 3.54 & 94.63 $\pm$ 2.8 & 88.01 $\pm$ 2.63 & 68.56 $\pm$ 1.14  \\ 
        \multirow{2}{*}{HoMM} & DEV & 59.02 $\pm$ 0.3 & 97.07 $\pm$ 1.76 & 68.85 $\pm$ 7.73 & 92.02 $\pm$ 6.98 & 86.18 $\pm$ 12.25 & 84.19 $\pm$ 27.39 & 79.61 $\pm$ 4.53 & 87.94 $\pm$ 5.1 & 83.51 $\pm$ 6.95 & 48.53 $\pm$ 6.31  \\ 
        ~ & FST & 63.62 $\pm$ 1.1 & 99.62 $\pm$ 0.67 & 83.81 $\pm$ 13.11 & 97.37 $\pm$ 4.55 & 98.26 $\pm$ 2.39 & 100.0 $\pm$ 0.0 & 83.75 $\pm$ 3.52 & 96.11 $\pm$ 0.55 & 91.4 $\pm$ 1.55 & 71.24 $\pm$ 8.73  \\ 
        ~ & TGT & 64.86 $\pm$ 2.01 & 99.62 $\pm$ 0.67 & 83.88 $\pm$ 12.29 & 97.37 $\pm$ 4.55 & 98.26 $\pm$ 2.39 & 100.0 $\pm$ 0.0 & 83.75 $\pm$ 3.52 & 95.25 $\pm$ 1.76 & 91.4 $\pm$ 1.55 & 70.8 $\pm$ 11.29  \\ \midrule

        ~ & SRC & 62.03 $\pm$ 2.37 & 99.33 $\pm$ 1.17 & 78.49 $\pm$ 4.95 & 80.4 $\pm$ 14.52 & 98.43 $\pm$ 1.92 & 92.75 $\pm$ 12.55 & 81.98 $\pm$ 2.06 & 96.45 $\pm$ 0.48 & 91.14 $\pm$ 3.8 & 66.19 $\pm$ 18.14  \\ 
        \multirow{2}{*}{MMDA} & DEV & 55.94 $\pm$ 1.24 & 100.0 $\pm$ 0.0 & 90.63 $\pm$ 0.8 & 84.76 $\pm$ 5.33 & 100.0 $\pm$ 0.0 & 100.0 $\pm$ 0.0 & 88.66 $\pm$ 0.0 & 96.73 $\pm$ 0.0 & 91.16 $\pm$ 3.76 & 76.99 $\pm$ 2.48 \\ 
        ~ & FST & 53.31 $\pm$ 3.46 & 98.08 $\pm$ 3.33 & 89.88 $\pm$ 3.04 & 82.43 $\pm$ 12.05 & 99.75 $\pm$ 0.43 & 100.0 $\pm$ 0.0 & 86.81 $\pm$ 3.21 & 96.73 $\pm$ 0.0 & 92.26 $\pm$ 1.86 & 80.37 $\pm$ 3.79   \\ 
        ~ & TGT & 60.98 $\pm$ 2.15 & 99.62 $\pm$ 0.67 & 90.91 $\pm$ 4.87 & 94.4 $\pm$ 7.01 & 99.51 $\pm$ 0.86 & 100.0 $\pm$ 0.0 & 88.66 $\pm$ 0.0 & 94.97 $\pm$ 3.04 & 90.47 $\pm$ 4.95 & 74.4 $\pm$ 3.27   \\ \midrule
        
        ~ & SRC & 66.98 $\pm$ 10.8 & 100.0 $\pm$ 0.0 & 82.73 $\pm$ 6.96 & 92.07 $\pm$ 10.97 & 92.86 $\pm$ 6.35 & 99.25 $\pm$ 0.65 & 86.05 $\pm$ 4.51 & 96.73 $\pm$ 0.0 & 92.35 $\pm$ 0.98 & 70.04 $\pm$ 2.97   \\ 
        \multirow{2}{*}{DSAN} & DEV & 65.87 $\pm$ 4.53 & 99.62 $\pm$ 0.67 & 82.92 $\pm$ 10.42 & 96.1 $\pm$ 6.75 & 92.05 $\pm$ 11.85 & 99.63 $\pm$ 0.65 & 88.13 $\pm$ 1.58 & 96.73 $\pm$ 0.0 & 92.65 $\pm$ 0.6 & 73.26 $\pm$ 4.76  \\ 
        ~ & FST & 70.72 $\pm$ 2.89 & 99.63 $\pm$ 0.65 & 81.75 $\pm$ 11.9 & 100.0 $\pm$ 0.0 & 99.5 $\pm$ 0.86 & 100.0 $\pm$ 0.0 & 92.72 $\pm$ 1.35 & 96.73 $\pm$ 0.0 & 92.97 $\pm$ 0.62 & 80.62 $\pm$ 2.13   \\ 
        ~ & TGT & 70.72 $\pm$ 2.89 & 99.63 $\pm$ 0.65 & 81.75 $\pm$ 11.9 & 100.0 $\pm$ 0.0 & 99.5 $\pm$ 0.86 & 100.0 $\pm$ 0.0 & 92.72 $\pm$ 1.35 & 96.73 $\pm$ 0.0 & 92.97 $\pm$ 0.62 & 80.62 $\pm$ 2.13  \\ \midrule

        ~ & SRC & 62.91 $\pm$ 0.48 & 100.0 $\pm$ 0.0 & 87.87 $\pm$ 3.14 & 79.67 $\pm$ 4.75 & 98.97 $\pm$ 1.2 & 100.0 $\pm$ 0.0 & 79.34 $\pm$ 2.88 & 94.63 $\pm$ 2.24 & 93.41 $\pm$ 0.12 & 76.31 $\pm$ 4.88  \\ 
        \multirow{2}{*}{DANN}& DEV & 60.39 $\pm$ 4.63 & 99.62 $\pm$ 0.67 & 70.43 $\pm$ 13.67 & 93.38 $\pm$ 6.22 & 90.11 $\pm$ 2.19 & 100.0 $\pm$ 0.0 & 79.53 $\pm$ 6.69 & 87.03 $\pm$ 10.72 & 91.89 $\pm$ 1.72 & 70.18 $\pm$ 7.36  \\ 
        ~ & FST & 60.69 $\pm$ 3.77 & 99.63 $\pm$ 0.65 & 84.39 $\pm$ 5.76 & 91.95 $\pm$ 10.9 & 99.01 $\pm$ 0.86 & 100.0 $\pm$ 0.0 & 77.91 $\pm$ 12.86 & 92.87 $\pm$ 2.45 & 93.01 $\pm$ 0.57 & 66.5 $\pm$ 8.34   \\ 
        ~ & TGT & 62.64 $\pm$ 0.82 & 100.0 $\pm$ 0.0 & 86.17 $\pm$ 3.44 & 87.54 $\pm$ 11.44 & 98.77 $\pm$ 1.13 & 100.0 $\pm$ 0.0 & 83.11 $\pm$ 4.81 & 94.63 $\pm$ 2.24 & 93.33 $\pm$ 0.0 & 76.41 $\pm$ 4.47 \\ \midrule

        ~ & SRC & 61.42 $\pm$ 2.1 & 100.0 $\pm$ 0.0 & 89.41 $\pm$ 5.14 & 98.89 $\pm$ 1.93 & 91.07 $\pm$ 12.3 & 100.0 $\pm$ 0.0 & 87.92 $\pm$ 1.28 & 83.62 $\pm$ 22.71 & 92.66 $\pm$ 1.17 & 91.4 $\pm$ 7.82   \\ 
        \multirow{2}{*}{DIRT-T} & DEV & 69.22 $\pm$ 4.27 & 100.0 $\pm$ 0.0 & 91.14 $\pm$ 7.55 & 100.0 $\pm$ 0.0 & 96.81 $\pm$ 4.3 & 100.0 $\pm$ 0.0 & 90.19 $\pm$ 2.66 & 96.73 $\pm$ 0.0 & 93.33 $\pm$ 0.0 & 86.03 $\pm$ 9.32   \\ 
        ~ & FST & 76.14 $\pm$ 3.86 & 100.0 $\pm$ 0.0 & 81.09 $\pm$ 5.81 & 100.0 $\pm$ 0.0 & 90.07 $\pm$ 11.43 & 100.0 $\pm$ 0.0 & 88.55 $\pm$ 2.73 & 96.73 $\pm$ 0.0 & 92.97 $\pm$ 0.62 & 82.33 $\pm$ 4.86   \\ 
        ~ & TGT & 80.81 $\pm$ 9.73 & 100.0 $\pm$ 0.0 & 84.02 $\pm$ 10.84 & 98.89 $\pm$ 1.93 & 99.25 $\pm$ 0.0 & 100.0 $\pm$ 0.0 & 89.04 $\pm$ 0.34 & 94.97 $\pm$ 3.04 & 93.73 $\pm$ 0.69 & 96.07 $\pm$ 0.56   \\\midrule

        ~ & SRC & 59.13 $\pm$ 13.61 & 100.0 $\pm$ 0.0 & 90.71 $\pm$ 3.37 & 72.95 $\pm$ 8.99 & 98.83 $\pm$ 0.74 & 100.0 $\pm$ 0.0 & 87.67 $\pm$ 1.72 & 89.17 $\pm$ 13.09 & 93.33 $\pm$ 0.0 & 67.96 $\pm$ 9.07  \\ 
        \multirow{2}{*}{CDAN} & DEV & 67.83 $\pm$ 2.1 & 99.66 $\pm$ 0.59 & 89.67 $\pm$ 7.27 & 84.69 $\pm$ 11.36 & 93.44 $\pm$ 0.87 & 85.13 $\pm$ 11.9 & 84.48 $\pm$ 4.41 & 89.41 $\pm$ 11.25 & 93.33 $\pm$ 0.0 & 53.35 $\pm$ 6.0  \\ 
        ~ & FST & 59.62 $\pm$ 2.71 & 99.62 $\pm$ 0.67 & 89.89 $\pm$ 4.93 & 95.2 $\pm$ 6.23 & 99.51 $\pm$ 0.86 & 100.0 $\pm$ 0.0 & 89.86 $\pm$ 1.5 & 92.61 $\pm$ 4.8 & 93.33 $\pm$ 0.0 & 69.08 $\pm$ 17.94  \\ 
        ~ & TGT & 61.59 $\pm$ 4.62 & 100.0 $\pm$ 0.0 & 86.81 $\pm$ 8.53 & 89.66 $\pm$ 8.98 & 99.75 $\pm$ 0.43 & 100.0 $\pm$ 0.0 & 88.66 $\pm$ 0.0 & 94.98 $\pm$ 3.03 & 93.33 $\pm$ 0.0 & 92.43 $\pm$ 2.8  \\ \midrule

        ~ & SRC & 60.81 $\pm$ 4.57 & 99.62 $\pm$ 0.67 & 68.77 $\pm$ 11.98 & 78.06 $\pm$ 19.52 & 95.77 $\pm$ 6.7 & 99.64 $\pm$ 0.62 & 87.78 $\pm$ 1.21 & 96.39 $\pm$ 0.59 & 92.97 $\pm$ 0.62 & 63.21 $\pm$ 1.74  \\ 
        \multirow{2}{*}{CoDATS} & DEV & 53.0 $\pm$ 11.34 & 91.12 $\pm$ 12.41 & 71.21 $\pm$ 3.99 & 81.77 $\pm$ 8.72 & 94.62 $\pm$ 0.39 & 93.65 $\pm$ 8.52 & 79.25 $\pm$ 16.94 & 85.87 $\pm$ 14.86 & 82.71 $\pm$ 8.85 & 53.74 $\pm$ 5.43  \\ 
        ~ & FST & 59.94 $\pm$ 3.33 & 99.23 $\pm$ 0.67 & 79.7 $\pm$ 11.44 & 87.75 $\pm$ 3.72 & 97.54 $\pm$ 1.83 & 98.18 $\pm$ 3.15 & 88.57 $\pm$ 0.87 & 92.17 $\pm$ 2.84 & 89.28 $\pm$ 4.07 & 77.54 $\pm$ 19.02   \\ 
        ~ & TGT & 62.52 $\pm$ 3.48 & 98.13 $\pm$ 2.31 & 86.64 $\pm$ 2.63 & 91.12 $\pm$ 11.2 & 97.47 $\pm$ 4.38 & 99.63 $\pm$ 0.64 & 81.87 $\pm$ 4.45 & 96.44 $\pm$ 0.5 & 92.61 $\pm$ 0.62 & 75.55 $\pm$ 14.82  \\ \midrule

        ~ & SRC & 59.88 $\pm$ 0.42 & 92.72 $\pm$ 1.91 & 59.57 $\pm$ 9.75 & 92.52 $\pm$ 4.16 & 81.79 $\pm$ 7.42 & 89.24 $\pm$ 13.96 & 80.27 $\pm$ 3.04 & 90.61 $\pm$ 4.02 & 88.61 $\pm$ 0.5 & 45.78 $\pm$ 9.6  \\ 
        \multirow{2}{*}{AdvSKM} & DEV & 58.65 $\pm$ 0.14 & 99.62 $\pm$ 0.67 & 65.99 $\pm$ 2.02 & 91.06 $\pm$ 4.51 & 97.61 $\pm$ 2.07 & 100.0 $\pm$ 0.0 & 79.87 $\pm$ 0.0 & 80.08 $\pm$ 4.84 & 85.63 $\pm$ 3.56 & 41.61 $\pm$ 8.84  \\ 
        ~ & FST & 59.23 $\pm$ 0.93 & 99.23 $\pm$ 1.33 & 65.99 $\pm$ 2.02 & 91.06 $\pm$ 4.51 & 97.36 $\pm$ 2.33 & 100.0 $\pm$ 0.0 & 79.87 $\pm$ 0.0 & 80.53 $\pm$ 4.8 & 85.25 $\pm$ 4.07 & 41.45 $\pm$ 8.41  \\ 
        ~ & TGT & 59.18 $\pm$ 0.98 & 99.62 $\pm$ 0.67 & 65.99 $\pm$ 2.02 & 91.06 $\pm$ 4.51 & 97.61 $\pm$ 2.07 & 100.0 $\pm$ 0.0 & 79.87 $\pm$ 0.0 & 80.08 $\pm$ 4.84 & 85.63 $\pm$ 3.56 & 41.94 $\pm$ 8.4 \\ \midrule

        ~ & SRC & 64.72 $\pm$ 6.44  &91.25 $\pm$ 8.01  &58.93 $\pm$ 7.37  &91.71 $\pm$ 3.87  &93.44 $\pm$ 4.09  &98.33 $\pm$ 0.0  &77.71 $\pm$ 4.31  &80.73 $\pm$ 8.87  &92.45 $\pm$ 0.65  &78.85 $\pm$ 4.29  \\ 
        \multirow{2}{*}{SASA} & DEV & 67.65 $\pm$ 7.41  &91.22 $\pm$ 11.81  &73.9 $\pm$ 7.81  &82.01 $\pm$ 16.19  &95.72 $\pm$ 3.9  &93.82 $\pm$ 9.59  &80.3 $\pm$ 5.21  &80.93 $\pm$ 14.55  &81.53 $\pm$ 7.81  &69.71 $\pm$ 8.23  \\ 
        ~ & FST & 61.9 $\pm$ 1.01  &90.94 $\pm$ 3.23  &65.75 $\pm$ 4.53  &95.62 $\pm$ 4.99  &92.14 $\pm$ 2.83  &99.44 $\pm$ 0.96  &73.88 $\pm$ 1.51  &90.02 $\pm$ 5.49  &92.11 $\pm$ 1.08  &78.42 $\pm$ 10.14  \\ 
        ~ & TGT & 60.18 $\pm$ 0.88  &94.2 $\pm$ 4.65  &71.79 $\pm$ 13.43  &94.87 $\pm$ 5.5  &97.73 $\pm$ 2.1  &98.89 $\pm$ 0.96  &78.88 $\pm$ 6.12  &85.22 $\pm$ 13.23  &91.68 $\pm$ 1.75  &76.53 $\pm$ 4.29   \\ \bottomrule

    \end{NiceTabular}}
\end{table}

\begin{table}[!ht]
    \centering
    \caption{Detailed results of 10 scenarios on WISDM dataset in terms of MF1 score.}
    \label{tbl:detailed_wisd}
    \resizebox{\textwidth}{!}{
    \begin{NiceTabular}{l|c|c|c|c|c|c|c|c|c|c|c}
    \toprule
    Algorithm &RISK & 7$\rightarrow$18 & 20$\rightarrow$30 & 35$\rightarrow$31 & 17$\rightarrow$23 & 6$\rightarrow$19 & 2$\rightarrow$111 & 33$\rightarrow$12 & 5$\rightarrow$26 & 28$\rightarrow$4 & 23$\rightarrow$32 \\ \midrule
    
    ~ & SRC & 29.39 $\pm$ 0.42  &64.62 $\pm$ 6.55  &42.08 $\pm$ 7.69  &81.89 $\pm$ 6.42  &66.7 $\pm$ 14.25  &50.55 $\pm$ 9.52  &47.93 $\pm$ 4.68  &29.71 $\pm$ 1.81  &56.79 $\pm$ 20.29  &49.2 $\pm$ 11.85  \\
    \multirow{2}{*}{DDC} & DEV & 29.27 $\pm$ 0.63  &63.96 $\pm$ 5.4  &41.54 $\pm$ 7.32  &81.89 $\pm$ 6.42  &59.71 $\pm$ 17.34  &51.34 $\pm$ 9.73  &46.11 $\pm$ 9.37  &29.71 $\pm$ 1.81  &57.44 $\pm$ 19.8  &47.62 $\pm$ 8.89  \\
    ~ & FST & 53.25 $\pm$ 1.55  &61.14 $\pm$ 0.91  &60.49 $\pm$ 1.95  &78.47 $\pm$ 1.43  &68.85 $\pm$ 10.14  &47.48 $\pm$ 3.63  &39.61 $\pm$ 12.87  &30.07 $\pm$ 1.36  &52.48 $\pm$ 8.51  &44.54 $\pm$ 4.6  \\
    ~ & TGT & 50.68 $\pm$ 1.32  &63.01 $\pm$ 1.06  &60.27 $\pm$ 2.17  &76.44 $\pm$ 1.14  &71.94 $\pm$ 8.66  &48.88 $\pm$ 4.85  &37.48 $\pm$ 13.57  &31.28 $\pm$ 2.85  &53.19 $\pm$ 7.64  &44.61 $\pm$ 5.63  \\ 
    
    \midrule
    
    ~ & SRC & 30.06 $\pm$ 0.37  &64.88 $\pm$ 5.15  &42.05 $\pm$ 8.59  &82.67 $\pm$ 5.87  &66.7 $\pm$ 14.25  &52.71 $\pm$ 9.06  &41.37 $\pm$ 11.16  &29.71 $\pm$ 1.81  &57.05 $\pm$ 19.85  &44.13 $\pm$ 10.12  \\
    \multirow{2}{*}{DCORAL} & DEV & 29.67 $\pm$ 0.09  &64.62 $\pm$ 4.9  &42.06 $\pm$ 8.51  &81.89 $\pm$ 6.42  &61.85 $\pm$ 15.48  &51.49 $\pm$ 8.99  &49.23 $\pm$ 6.53  &29.84 $\pm$ 1.71  &57.83 $\pm$ 20.46  &46.92 $\pm$ 9.76  \\
    ~ & FST & 42.54 $\pm$ 11.29  &63.3 $\pm$ 2.65  &55.27 $\pm$ 16.54  &81.38 $\pm$ 1.76  &70.64 $\pm$ 11.13  &49.54 $\pm$ 4.66  &42.44 $\pm$ 15.15  &29.6 $\pm$ 2.12  &52.5 $\pm$ 8.46  &44.76 $\pm$ 5.27  \\
    ~ & TGT & 50.68 $\pm$ 1.32  &64.63 $\pm$ 3.07  &60.29 $\pm$ 2.14  &76.94 $\pm$ 1.0  &73.65 $\pm$ 6.13  &49.21 $\pm$ 4.99  &37.58 $\pm$ 13.42  &31.28 $\pm$ 2.85  &53.22 $\pm$ 7.57  &44.45 $\pm$ 5.75  \\
     
    \midrule
    
    ~ & SRC & 26.04 $\pm$ 1.4  &80.53 $\pm$ 8.13  &45.66 $\pm$ 6.89  &78.65 $\pm$ 2.49  &65.12 $\pm$ 9.43  &43.37 $\pm$ 1.97  &55.12 $\pm$ 10.08  &28.85 $\pm$ 1.57  &51.64 $\pm$ 4.96  &44.77 $\pm$ 1.91  \\
    \multirow{2}{*}{HoMM} & DEV & 27.28 $\pm$ 2.08  &81.82 $\pm$ 7.59  &42.34 $\pm$ 7.28  &83.8 $\pm$ 4.75  &67.8 $\pm$ 20.12  &51.15 $\pm$ 7.01  &56.83 $\pm$ 7.36  &27.93 $\pm$ 0.88  &54.37 $\pm$ 8.36  &48.86 $\pm$ 7.5  \\
    ~ & FST & 47.95 $\pm$ 8.04  &61.73 $\pm$ 2.78  &56.19 $\pm$ 3.5  &81.99 $\pm$ 1.84  &59.92 $\pm$ 15.49  &48.04 $\pm$ 2.81  &58.16 $\pm$ 9.24  &30.52 $\pm$ 1.93  &53.57 $\pm$ 7.17  &42.46 $\pm$ 3.66  \\
    ~ & TGT & 45.53 $\pm$ 4.24  &83.65 $\pm$ 5.51  &50.22 $\pm$ 3.82  &84.15 $\pm$ 2.49  &57.0 $\pm$ 15.97  &54.34 $\pm$ 11.85  &60.44 $\pm$ 12.44  &27.03 $\pm$ 3.65  &50.56 $\pm$ 4.4  &56.27 $\pm$ 1.92  \\
    
    \midrule
    
    ~ & SRC & 36.28 $\pm$ 1.99  &64.29 $\pm$ 1.27  &76.35 $\pm$ 7.65  &75.95 $\pm$ 1.33  &76.03 $\pm$ 1.61  &45.98 $\pm$ 1.66  &56.74 $\pm$ 8.98  &28.26 $\pm$ 0.54  &64.8 $\pm$ 5.52  &47.73 $\pm$ 1.25  \\
    \multirow{2}{*}{MMDA} & DEV & 38.5 $\pm$ 1.42  &57.49 $\pm$ 1.17  &83.09 $\pm$ 1.78  &76.02 $\pm$ 0.2  &62.24 $\pm$ 12.51  &50.99 $\pm$ 10.09  &56.22 $\pm$ 8.08  &30.08 $\pm$ 2.59  &65.64 $\pm$ 8.71  &48.34 $\pm$ 1.03  \\
    ~ & FST & 50.83 $\pm$ 1.55  &62.94 $\pm$ 1.71  &61.07 $\pm$ 2.94  &80.87 $\pm$ 0.77  &76.07 $\pm$ 3.4  &48.12 $\pm$ 3.24  &46.43 $\pm$ 4.17  &28.65 $\pm$ 0.47  &53.41 $\pm$ 8.05  &43.46 $\pm$ 4.94  \\
    ~ & TGT & 55.87 $\pm$ 12.05  &62.29 $\pm$ 5.58  &84.19 $\pm$ 0.45  &82.31 $\pm$ 4.1  &65.6 $\pm$ 8.02  &42.2 $\pm$ 3.77  &67.02 $\pm$ 3.26  &28.85 $\pm$ 2.61  &59.18 $\pm$ 6.85  &50.66 $\pm$ 1.44  \\
    
    \midrule

    ~ & SRC & 44.29 $\pm$ 8.32  &78.21 $\pm$ 2.57  &49.08 $\pm$ 2.76  &81.82 $\pm$ 9.65  &62.11 $\pm$ 12.23  &69.66 $\pm$ 7.7  &66.7 $\pm$ 2.81  &29.2 $\pm$ 0.85  &57.22 $\pm$ 4.59  &51.56 $\pm$ 4.37  \\
    \multirow{2}{*}{DSAN} & DEV & 49.47 $\pm$ 2.58  &80.26 $\pm$ 1.87  &52.23 $\pm$ 0.97  &80.72 $\pm$ 5.99  &61.85 $\pm$ 11.4  &69.48 $\pm$ 7.39  &65.48 $\pm$ 3.47  &29.42 $\pm$ 0.89  &62.21 $\pm$ 6.01  &50.12 $\pm$ 3.22  \\
    ~ & FST & 51.74 $\pm$ 1.35  &64.33 $\pm$ 1.89  &56.51 $\pm$ 19.7  &81.52 $\pm$ 1.98  &76.29 $\pm$ 3.74  &48.89 $\pm$ 3.48  &51.98 $\pm$ 1.35  &29.93 $\pm$ 1.43  &52.73 $\pm$ 7.59  &57.08 $\pm$ 5.23  \\
    ~ & TGT & 60.46 $\pm$ 3.34  &66.03 $\pm$ 4.11  &53.15 $\pm$ 1.14  &89.83 $\pm$ 11.88  &71.12 $\pm$ 6.41  &63.93 $\pm$ 12.32  &65.92 $\pm$ 2.31  &41.84 $\pm$ 9.83  &56.81 $\pm$ 17.64  &41.68 $\pm$ 2.11  \\
    
    \midrule

    ~ & SRC & 54.74 $\pm$ 10.21  &56.36 $\pm$ 1.55  &51.11 $\pm$ 11.18  &81.54 $\pm$ 11.35  &73.83 $\pm$ 2.27  &62.75 $\pm$ 11.15  &44.67 $\pm$ 11.22  &36.88 $\pm$ 8.81  &63.28 $\pm$ 14.61  &39.75 $\pm$ 9.77  \\
    \multirow{2}{*}{DANN} & DEV & 59.78 $\pm$ 17.81  &60.06 $\pm$ 5.93  &63.14 $\pm$ 4.73  &84.83 $\pm$ 9.13  &77.38 $\pm$ 3.54  &69.6 $\pm$ 10.82  &60.96 $\pm$ 7.99  &36.55 $\pm$ 7.7  &63.21 $\pm$ 14.3  &48.55 $\pm$ 2.18  \\
    ~ & FST & 26.43 $\pm$ 2.73  &59.0 $\pm$ 4.83  &38.31 $\pm$ 18.38  &76.57 $\pm$ 9.74  &56.8 $\pm$ 2.06  &42.79 $\pm$ 1.0  &36.22 $\pm$ 16.46  &29.21 $\pm$ 0.35  &48.61 $\pm$ 14.4  &49.94 $\pm$ 8.62  \\
    ~ & TGT & 59.78 $\pm$ 17.81  &60.06 $\pm$ 5.93  &63.14 $\pm$ 4.73  &84.83 $\pm$ 9.13  &77.38 $\pm$ 3.54  &69.6 $\pm$ 10.82  &60.96 $\pm$ 7.99  &36.55 $\pm$ 7.7  &63.21 $\pm$ 14.3  &48.55 $\pm$ 2.18  \\
    
    \midrule
    
    ~ & SRC & 40.31 $\pm$ 0.74  &45.6 $\pm$ 5.55  &29.8 $\pm$ 12.21  &63.71 $\pm$ 0.81  &29.41 $\pm$ 9.06  &33.01 $\pm$ 8.85  &40.93 $\pm$ 4.8  &27.07 $\pm$ 0.51  &46.67 $\pm$ 12.73  &56.27 $\pm$ 9.98  \\
    \multirow{2}{*}{CDAN} & DEV & 36.21 $\pm$ 9.52  &64.95 $\pm$ 0.78  &7.51 $\pm$ 5.65  &77.58 $\pm$ 1.05  &52.25 $\pm$ 9.96  &42.96 $\pm$ 1.25  &52.83 $\pm$ 16.08  &28.14 $\pm$ 0.66  &44.65 $\pm$ 0.88  &57.54 $\pm$ 0.54  \\
    ~ & FST & 30.04 $\pm$ 5.2  &59.64 $\pm$ 3.58  &46.83 $\pm$ 8.89  &86.15 $\pm$ 14.23  &73.51 $\pm$ 9.8  &64.77 $\pm$ 5.65  &59.33 $\pm$ 10.26  &28.98 $\pm$ 1.4  &54.65 $\pm$ 17.72  &51.99 $\pm$ 16.42  \\
    ~ & TGT & 30.04 $\pm$ 5.2  &59.64 $\pm$ 3.58  &46.83 $\pm$ 8.89  &86.15 $\pm$ 14.23  &73.51 $\pm$ 9.8  &64.77 $\pm$ 5.65  &59.33 $\pm$ 10.26  &28.98 $\pm$ 1.4  &54.65 $\pm$ 17.72  &51.99 $\pm$ 16.42  \\
    
    \midrule
    
    ~ & SRC & 33.53 $\pm$ 9.56  &62.85 $\pm$ 0.83  &52.16 $\pm$ 13.65  &83.24 $\pm$ 8.63  &55.15 $\pm$ 3.99  &63.18 $\pm$ 18.79  &48.09 $\pm$ 20.05  &28.09 $\pm$ 0.26  &54.53 $\pm$ 8.97  &51.45 $\pm$ 8.71  \\
    \multirow{2}{*}{DIRT-T} & DEV & 33.53 $\pm$ 9.56  &62.85 $\pm$ 0.83  &52.16 $\pm$ 13.65  &83.24 $\pm$ 8.63  &55.15 $\pm$ 3.99  &63.18 $\pm$ 18.79  &48.09 $\pm$ 20.05  &28.09 $\pm$ 0.26  &54.53 $\pm$ 8.97  &51.45 $\pm$ 8.71  \\
    ~ & FST & 47.16 $\pm$ 16.4  &62.01 $\pm$ 7.77  &51.43 $\pm$ 12.03  &85.86 $\pm$ 9.87  &47.7 $\pm$ 13.68  &48.54 $\pm$ 10.08  &63.14 $\pm$ 12.88  &50.24 $\pm$ 8.61  &55.43 $\pm$ 9.02  &49.08 $\pm$ 8.03  \\
    ~ & TGT & 47.0 $\pm$ 6.15  &60.53 $\pm$ 0.39  &56.57 $\pm$ 0.0  &78.26 $\pm$ 3.09  &54.14 $\pm$ 11.37  &72.57 $\pm$ 9.62  &77.99 $\pm$ 2.21  &37.54 $\pm$ 8.15  &59.71 $\pm$ 2.11  &51.59 $\pm$ 4.6  \\
    
    \midrule
     
    ~ & SRC & 29.68 $\pm$ 4.82  &69.03 $\pm$ 6.83  &44.46 $\pm$ 12.1  &75.84 $\pm$ 1.24  &72.43 $\pm$ 2.57  &42.49 $\pm$ 6.39  &51.17 $\pm$ 13.15  &29.92 $\pm$ 1.89  &53.72 $\pm$ 6.45  &41.06 $\pm$ 5.51  \\
    \multirow{2}{*}{AdvSKM} & DEV & 22.31 $\pm$ 2.65  &67.27 $\pm$ 0.62  &49.45 $\pm$ 12.41  &76.52 $\pm$ 9.07  &68.45 $\pm$ 5.55  &53.2 $\pm$ 9.87  &49.9 $\pm$ 18.16  &29.59 $\pm$ 1.58  &51.48 $\pm$ 3.87  &39.47 $\pm$ 0.57  \\
    ~ & FST & 44.96 $\pm$ 14.46  &65.2 $\pm$ 3.86  &58.6 $\pm$ 0.63  &72.68 $\pm$ 3.18  &72.3 $\pm$ 0.5  &45.9 $\pm$ 4.18  &50.84 $\pm$ 15.81  &30.5 $\pm$ 1.78  &51.39 $\pm$ 3.3  &44.85 $\pm$ 2.09  \\
    ~ & TGT & 50.29 $\pm$ 3.46  &63.98 $\pm$ 4.67  &58.58 $\pm$ 0.63  &72.11 $\pm$ 2.42  &72.25 $\pm$ 0.59  &47.02 $\pm$ 6.12  &54.42 $\pm$ 14.83  &31.07 $\pm$ 1.38  &55.09 $\pm$ 8.61  &43.43 $\pm$ 3.49  \\
    
    \midrule

    ~ & SRC & 35.11 $\pm$ 11.44  &60.8 $\pm$ 1.13  &31.57 $\pm$ 12.59  &87.72 $\pm$ 5.35  &47.67 $\pm$ 5.89  &72.39 $\pm$ 8.58  &35.66 $\pm$ 7.98  &29.66 $\pm$ 0.56  &47.39 $\pm$ 11.91  &29.89 $\pm$ 7.15  \\
    \multirow{2}{*}{SASA} & DEV & 14.66 $\pm$ 7.12  &9.15 $\pm$ 0.0  &3.07 $\pm$ 0.0  &8.84 $\pm$ 2.31  &8.25 $\pm$ 0.0  &8.94 $\pm$ 0.68  &8.63 $\pm$ 0.0  &9.86 $\pm$ 2.0  &6.34 $\pm$ 0.0  &9.3 $\pm$ 0.0  \\
    ~ & FST & 27.52 $\pm$ 6.51  &69.0 $\pm$ 5.58  &39.74 $\pm$ 17.8  &77.43 $\pm$ 5.6  &48.2 $\pm$ 4.63  &49.32 $\pm$ 10.68  &38.23 $\pm$ 18.86  &31.45 $\pm$ 2.85  &44.61 $\pm$ 17.34  &37.47 $\pm$ 5.47  \\
    ~ & TGT & 38.16 $\pm$ 8.74  &68.44 $\pm$ 0.21  &41.66 $\pm$ 2.68  &83.77 $\pm$ 5.42  &56.32 $\pm$ 7.42  &75.76 $\pm$ 13.74  &51.04 $\pm$ 8.53  &28.46 $\pm$ 0.59  &51.03 $\pm$ 14.35  &38.75 $\pm$ 4.82  \\

    \bottomrule

    \end{NiceTabular}}
\end{table}
\begin{table}[!ht]
    \centering
    \caption{Detailed results of 10 scenarios on EEG dataset in terms of MF1 score.}
    \resizebox{\textwidth}{!}{
    \begin{NiceTabular}{l|c|c|c|c|c|c|c|c|c|c|c}
    \toprule
    Algorithm &RISK & 0$\rightarrow$11 & 7$\rightarrow$18 & 9$\rightarrow$14 & 12$\rightarrow$5 & 16$\rightarrow$1 & 3$\rightarrow$19 & 18$\rightarrow$12 & 13$\rightarrow$17 & 5$\rightarrow$15 & 6$\rightarrow$2  \\ \midrule

    ~ & SRC & 54.73 $\pm$ 3.9  &52.01 $\pm$ 5.13  &51.67 $\pm$ 3.45  &56.69 $\pm$ 4.92  &54.32 $\pm$ 4.19  &58.92 $\pm$ 10.43  &72.98 $\pm$ 3.6  &71.03 $\pm$ 1.68  &67.91 $\pm$ 1.98  &67.9 $\pm$ 6.31  \\
    \multirow{2}{*}{DDC} & DEV & 54.73 $\pm$ 3.94  &52.19 $\pm$ 5.03  &51.67 $\pm$ 3.45  &56.71 $\pm$ 5.07  &54.29 $\pm$ 4.21  &58.92 $\pm$ 10.3  &73.11 $\pm$ 3.44  &70.97 $\pm$ 1.76  &67.91 $\pm$ 1.98  &67.92 $\pm$ 6.22  \\
    ~ & FST & 54.69 $\pm$ 3.89  &52.34 $\pm$ 4.9  &51.64 $\pm$ 3.45  &56.79 $\pm$ 4.94  &54.17 $\pm$ 4.17  &58.92 $\pm$ 10.42  &73.18 $\pm$ 3.37  &70.96 $\pm$ 1.82  &67.91 $\pm$ 1.98  &67.99 $\pm$ 6.25  \\
    ~ & TGT & 54.7 $\pm$ 3.88  &52.02 $\pm$ 4.52  &51.76 $\pm$ 3.58  &56.75 $\pm$ 4.9  &54.5 $\pm$ 4.33  &58.77 $\pm$ 10.69  &73.28 $\pm$ 3.27  &70.98 $\pm$ 1.83  &67.95 $\pm$ 2.04  &68.06 $\pm$ 6.27 \\ \midrule
     
    ~ & SRC & 53.04 $\pm$ 3.02  &52.01 $\pm$ 5.2  &51.64 $\pm$ 3.45  &56.67 $\pm$ 5.03  &54.03 $\pm$ 4.16  &59.14 $\pm$ 10.31  &73.25 $\pm$ 3.24  &70.88 $\pm$ 1.7  &67.95 $\pm$ 2.04  &67.8 $\pm$ 6.26  \\
    \multirow{2}{*}{DCORAL} & DEV & 46.63 $\pm$ 3.15  &55.44 $\pm$ 6.53  &40.26 $\pm$ 12.78  &53.66 $\pm$ 8.1  &52.87 $\pm$ 11.4  &53.21 $\pm$ 10.22  &61.62 $\pm$ 6.74  &73.0 $\pm$ 3.58  &61.67 $\pm$ 6.46  &63.2 $\pm$ 13.61  \\
    ~ & FST & 54.78 $\pm$ 3.44  &52.05 $\pm$ 5.08  &51.7 $\pm$ 3.5  &56.67 $\pm$ 5.03  &54.28 $\pm$ 4.22  &58.87 $\pm$ 10.49  &73.12 $\pm$ 3.34  &70.99 $\pm$ 1.89  &67.95 $\pm$ 2.04  &68.03 $\pm$ 6.15  \\
    ~ & TGT & 51.03 $\pm$ 2.52  &55.06 $\pm$ 3.0  &50.57 $\pm$ 3.87  &56.86 $\pm$ 3.8  &53.22 $\pm$ 3.53  &60.4 $\pm$ 8.69  &75.82 $\pm$ 0.18  &70.65 $\pm$ 2.07  &69.59 $\pm$ 1.53  &67.24 $\pm$ 5.42 \\ \midrule
    
    ~ & SRC & 52.06 $\pm$ 4.34  &52.26 $\pm$ 4.9  &51.62 $\pm$ 3.46  &56.65 $\pm$ 5.0  &54.23 $\pm$ 3.96  &59.13 $\pm$ 10.23  &73.24 $\pm$ 3.23  &71.09 $\pm$ 1.77  &67.95 $\pm$ 2.04  &67.77 $\pm$ 6.29  \\
    \multirow{2}{*}{HoMM} & DEV & 45.68 $\pm$ 7.01  &43.28 $\pm$ 20.24  &39.26 $\pm$ 5.91  &51.76 $\pm$ 10.5  &44.67 $\pm$ 18.91  &56.86 $\pm$ 6.57  &58.37 $\pm$ 11.19  &69.47 $\pm$ 5.85  &66.57 $\pm$ 1.9  &69.31 $\pm$ 2.97  \\
    ~ & FST & 51.58 $\pm$ 4.41  &52.53 $\pm$ 4.66  &51.71 $\pm$ 3.44  &56.67 $\pm$ 5.02  &54.22 $\pm$ 4.13  &59.29 $\pm$ 10.18  &73.28 $\pm$ 3.24  &70.97 $\pm$ 1.61  &67.99 $\pm$ 2.1  &67.86 $\pm$ 6.54 \\
    ~ & TGT & 49.26 $\pm$ 6.57  &53.66 $\pm$ 4.98  &48.83 $\pm$ 7.29  &58.76 $\pm$ 3.88  &53.89 $\pm$ 6.4  &65.86 $\pm$ 7.99  &70.49 $\pm$ 1.6  &72.78 $\pm$ 1.91  &67.72 $\pm$ 1.42  &66.84 $\pm$ 10.3  \\ \midrule

    ~ & SRC & 47.2 $\pm$ 5.55  &53.23 $\pm$ 4.85  &51.78 $\pm$ 3.42  &57.16 $\pm$ 4.76  &54.23 $\pm$ 3.74  &60.21 $\pm$ 9.22  &74.84 $\pm$ 1.38  &71.03 $\pm$ 1.65  &68.23 $\pm$ 1.89  &68.09 $\pm$ 6.04  \\
    \multirow{2}{*}{MMDA} & DEV & 38.1 $\pm$ 7.13  &63.97 $\pm$ 5.81  &47.4 $\pm$ 3.9  &53.21 $\pm$ 7.86  &50.03 $\pm$ 5.26  &62.73 $\pm$ 15.12  &68.82 $\pm$ 5.54  &69.74 $\pm$ 6.05  &66.6 $\pm$ 1.72  &65.7 $\pm$ 6.47  \\
    ~ & FST & 32.99 $\pm$ 4.76  &65.74 $\pm$ 1.71  &62.16 $\pm$ 2.44  &56.21 $\pm$ 1.18  &53.07 $\pm$ 2.65  &69.15 $\pm$ 1.13  &71.58 $\pm$ 1.35  &65.6 $\pm$ 1.59  &70.75 $\pm$ 2.56  &64.18 $\pm$ 1.71  \\
    ~ & TGT & 45.48 $\pm$ 2.71  &64.14 $\pm$ 2.17  &57.61 $\pm$ 2.09  &57.77 $\pm$ 0.99  &54.7 $\pm$ 3.05  &71.43 $\pm$ 0.88  &75.37 $\pm$ 1.46  &71.4 $\pm$ 2.0  &70.61 $\pm$ 0.91  &66.18 $\pm$ 3.13  \\ \midrule
     
    ~ & SRC & 40.71 $\pm$ 0.3  &65.73 $\pm$ 4.03  &41.7 $\pm$ 8.2  &59.1 $\pm$ 3.4  &49.42 $\pm$ 6.35  &56.8 $\pm$ 7.73  &62.61 $\pm$ 5.49  &68.3 $\pm$ 4.97  &65.67 $\pm$ 1.93  &69.05 $\pm$ 3.53  \\
    \multirow{2}{*}{DSAN} & DEV & 38.09 $\pm$ 1.37  &66.65 $\pm$ 2.29  &46.64 $\pm$ 3.4  &55.2 $\pm$ 8.32  &49.83 $\pm$ 4.25  &61.56 $\pm$ 6.26  &67.11 $\pm$ 1.53  &70.8 $\pm$ 1.17  &70.13 $\pm$ 1.8  &67.7 $\pm$ 0.85  \\
    ~ & FST & 37.35 $\pm$ 1.45  &64.92 $\pm$ 1.97  &46.33 $\pm$ 2.87  &54.88 $\pm$ 1.73  &49.94 $\pm$ 4.46  &61.83 $\pm$ 7.23  &65.96 $\pm$ 2.11  &67.73 $\pm$ 0.74  &68.76 $\pm$ 2.65  &68.28 $\pm$ 3.03   \\
    ~ & TGT & 38.24 $\pm$ 0.96  &67.12 $\pm$ 2.46  &46.42 $\pm$ 3.88  &57.06 $\pm$ 3.04  &49.79 $\pm$ 3.43  &60.01 $\pm$ 7.21  &67.46 $\pm$ 0.99  &70.72 $\pm$ 1.86  &69.98 $\pm$ 1.75  &68.27 $\pm$ 3.23  \\ \midrule

    ~ & SRC & 52.63 $\pm$ 0.84  &55.87 $\pm$ 4.6  &51.25 $\pm$ 4.94  &55.31 $\pm$ 6.86  &52.64 $\pm$ 2.26  &58.46 $\pm$ 4.96  &71.71 $\pm$ 4.75  &71.83 $\pm$ 2.61  &70.3 $\pm$ 1.72  &68.01 $\pm$ 1.16  \\
    \multirow{2}{*}{DANN}  & DEV & 52.63 $\pm$ 0.84  &55.87 $\pm$ 4.6  &51.25 $\pm$ 4.94  &55.31 $\pm$ 6.86  &52.64 $\pm$ 2.26  &58.46 $\pm$ 4.96  &71.71 $\pm$ 4.75  &71.83 $\pm$ 2.61  &70.3 $\pm$ 1.72  &68.01 $\pm$ 1.16  \\
    ~ & FST & 52.63 $\pm$ 0.84  &55.87 $\pm$ 4.6  &51.25 $\pm$ 4.94  &55.31 $\pm$ 6.86  &52.64 $\pm$ 2.26  &58.46 $\pm$ 4.96  &71.71 $\pm$ 4.75  &71.83 $\pm$ 2.61  &70.3 $\pm$ 1.72  &68.01 $\pm$ 1.16 \\
    ~ & TGT & 52.63 $\pm$ 0.84  &55.87 $\pm$ 4.6  &51.25 $\pm$ 4.94  &55.31 $\pm$ 6.86  &52.64 $\pm$ 2.26  &58.46 $\pm$ 4.96  &71.71 $\pm$ 4.75  &71.83 $\pm$ 2.61  &70.3 $\pm$ 1.72  &68.01 $\pm$ 1.16  \\ \midrule

    ~ & SRC & 36.89 $\pm$ 0.81  &62.98 $\pm$ 4.57  &48.53 $\pm$ 4.4  &51.56 $\pm$ 6.56  &33.87 $\pm$ 6.32  &61.39 $\pm$ 6.84  &70.74 $\pm$ 2.52  &73.3 $\pm$ 1.69  &68.78 $\pm$ 1.43  &68.33 $\pm$ 2.39  \\
    \multirow{2}{*}{CDAN} & DEV & 36.8 $\pm$ 6.51  &62.91 $\pm$ 6.29  &40.34 $\pm$ 0.51  &58.58 $\pm$ 2.44  &36.09 $\pm$ 4.54  &62.84 $\pm$ 4.46  &72.92 $\pm$ 1.73  &73.09 $\pm$ 0.66  &71.06 $\pm$ 1.61  &65.31 $\pm$ 3.77  \\
    ~ & FST & 31.04 $\pm$ 3.95  &59.78 $\pm$ 5.06  &34.94 $\pm$ 8.11  &52.13 $\pm$ 4.0  &36.37 $\pm$ 0.99  &58.92 $\pm$ 11.16  &59.04 $\pm$ 6.81  &70.57 $\pm$ 1.84  &67.68 $\pm$ 0.5  &69.0 $\pm$ 2.76   \\
    ~ & TGT & 32.98 $\pm$ 1.57  &60.28 $\pm$ 4.06  &48.54 $\pm$ 4.99  &56.8 $\pm$ 4.48  &51.11 $\pm$ 1.9  &61.4 $\pm$ 2.25  &74.96 $\pm$ 1.09  &71.73 $\pm$ 1.31  &71.55 $\pm$ 0.53  &65.76 $\pm$ 5.32  \\  \midrule

    ~ & SRC & 37.21 $\pm$ 3.11  &61.92 $\pm$ 8.1  &49.34 $\pm$ 6.3  &60.8 $\pm$ 2.32  &41.9 $\pm$ 5.33  &64.72 $\pm$ 11.07  &70.64 $\pm$ 0.51  &73.54 $\pm$ 0.97  &73.19 $\pm$ 3.23  &74.91 $\pm$ 2.3  \\
    \multirow{2}{*}{DIRT-T} & DEV & 19.85 $\pm$ 7.62  &58.75 $\pm$ 9.06  &49.25 $\pm$ 1.76  &62.3 $\pm$ 1.98  &37.72 $\pm$ 0.62  &56.38 $\pm$ 5.11  &51.35 $\pm$ 0.28  &49.13 $\pm$ 1.59  &71.97 $\pm$ 0.76  &68.2 $\pm$ 3.95 \\
    ~ & FST & 34.35 $\pm$ 2.12  &66.21 $\pm$ 2.38  &48.64 $\pm$ 4.38  &61.08 $\pm$ 1.89  &38.37 $\pm$ 2.01  &66.15 $\pm$ 2.01  &69.38 $\pm$ 5.73  &72.44 $\pm$ 1.56  &68.96 $\pm$ 0.88  &68.4 $\pm$ 2.24  \\
    ~ & TGT & 28.93 $\pm$ 3.68  &62.41 $\pm$ 8.76  &46.16 $\pm$ 2.47  &61.78 $\pm$ 1.52  &57.35 $\pm$ 8.26  &68.56 $\pm$ 6.02  &68.87 $\pm$ 1.05  &73.68 $\pm$ 2.54  &73.45 $\pm$ 0.98  &72.66 $\pm$ 1.38  \\ \midrule

    ~ & SRC & 49.51 $\pm$ 1.21  &47.37 $\pm$ 9.89  &43.31 $\pm$ 2.57  &59.46 $\pm$ 3.97  &56.72 $\pm$ 5.45  &56.38 $\pm$ 3.76  &64.48 $\pm$ 9.91  &68.77 $\pm$ 3.04  &60.96 $\pm$ 3.54  &53.06 $\pm$ 10.48  \\
    \multirow{2}{*}{CoDATS} & DEV & 34.94 $\pm$ 10.37  &51.98 $\pm$ 12.08  &45.3 $\pm$ 0.55  &53.13 $\pm$ 7.15  &43.27 $\pm$ 5.48  &51.68 $\pm$ 13.12  &66.65 $\pm$ 2.05  &70.31 $\pm$ 5.43  &64.72 $\pm$ 0.84  &64.09 $\pm$ 2.73  \\
    ~ & FST & 33.69 $\pm$ 10.17  &50.77 $\pm$ 7.43  &49.85 $\pm$ 2.85  &60.52 $\pm$ 1.11  &48.33 $\pm$ 2.89  &55.49 $\pm$ 9.23  &67.09 $\pm$ 2.21  &68.05 $\pm$ 4.76  &63.44 $\pm$ 4.13  &59.21 $\pm$ 3.56 \\
    ~ & TGT & 35.18 $\pm$ 3.33  &54.19 $\pm$ 12.36  &37.6 $\pm$ 12.3  &56.32 $\pm$ 7.79  &50.71 $\pm$ 3.87  &64.36 $\pm$ 4.8  &73.53 $\pm$ 0.93  &71.19 $\pm$ 3.29  &64.12 $\pm$ 3.65  &66.02 $\pm$ 3.26  \\ \midrule
    
    ~ & SRC & 51.25 $\pm$ 1.45  &58.68 $\pm$ 4.69  &51.67 $\pm$ 0.79  &58.46 $\pm$ 0.77  &54.87 $\pm$ 2.11  &64.82 $\pm$ 2.19  &74.01 $\pm$ 0.7  &68.95 $\pm$ 2.33  &66.87 $\pm$ 1.34  &62.16 $\pm$ 5.26  \\
    \multirow{2}{*}{AdvSKM}  & DEV & 44.89 $\pm$ 3.73  &48.5 $\pm$ 18.98  &32.48 $\pm$ 15.44  &47.93 $\pm$ 10.85  &52.12 $\pm$ 7.71  &66.68 $\pm$ 4.51  &70.33 $\pm$ 3.74  &70.92 $\pm$ 2.71  &59.89 $\pm$ 3.62  &68.13 $\pm$ 3.98  \\
    ~ & FST & 50.62 $\pm$ 1.09  &58.9 $\pm$ 4.53  &51.7 $\pm$ 0.72  &58.56 $\pm$ 0.78  &54.93 $\pm$ 2.0  &64.53 $\pm$ 1.68  &74.15 $\pm$ 0.55  &68.93 $\pm$ 2.66  &66.91 $\pm$ 1.28  &61.79 $\pm$ 5.56 \\
    ~ & TGT & 51.25 $\pm$ 1.45  &58.68 $\pm$ 4.69  &51.67 $\pm$ 0.79  &58.46 $\pm$ 0.77  &54.87 $\pm$ 2.11  &64.82 $\pm$ 2.19  &74.01 $\pm$ 0.7  &68.95 $\pm$ 2.33  &66.87 $\pm$ 1.34  &62.16 $\pm$ 5.26   \\ \midrule

    ~ & SRC & 43.01 $\pm$ 1.44  &60.49 $\pm$ 2.95  &47.51 $\pm$ 4.81  &61.5 $\pm$ 4.09  &43.59 $\pm$ 3.3  &59.4 $\pm$ 0.61  &67.08 $\pm$ 1.89  &64.32 $\pm$ 4.71  &66.0 $\pm$ 0.74  &66.49 $\pm$ 9.61 \\ 
    \multirow{2}{*}{SASA}& DEV & 43.01 $\pm$ 1.44  & 60.49 $\pm$ 2.95  &47.51 $\pm$ 4.81  &61.5 $\pm$ 4.09  &43.59 $\pm$ 3.3  &59.4 $\pm$ 0.61  &67.08 $\pm$ 1.89  &64.32 $\pm$ 4.71  &66.0 $\pm$ 0.74  &66.49 $\pm$ 9.61 \\
    ~ & FST & 41.04 $\pm$ 3.69  &57.41 $\pm$ 2.65  &42.72 $\pm$ 1.97  &62.43 $\pm$ 2.06  &51.77 $\pm$ 3.04  &57.55 $\pm$ 2.11  &62.17 $\pm$ 9.01  &65.94 $\pm$ 1.41  &60.71 $\pm$ 2.93  &61.95 $\pm$ 1.63 \\
    ~ & TGT & 41.77 $\pm$ 4.84  &59.8 $\pm$ 6.0  &52.29 $\pm$ 1.22  &60.17 $\pm$ 2.9  &55.55 $\pm$ 2.79  &59.45 $\pm$ 6.84  &64.59 $\pm$ 11.58  &67.31 $\pm$ 1.33  &67.27 $\pm$ 1.14  &69.96 $\pm$ 6.21 \\

    \bottomrule
    \end{NiceTabular}}
    \label{tbl:detailed_eeg}
\end{table}

\begin{table}[!ht]
    \centering
    \caption{Detailed results of 10 scenarios on HHAR dataset in terms of MF1 score.}
    \label{tbl:detailed_HHAR}
    \resizebox{\textwidth}{!}{
    \begin{NiceTabular}{l|c|c|c|c|c|c|c|c|c|c|c}
    \toprule
        Algorithm & RISK & 0$\rightarrow$6 & 1$\rightarrow$6 & 2$\rightarrow$7 & 3$\rightarrow$8 & 4$\rightarrow$5 & 5$\rightarrow$0 & 6$\rightarrow$1 & 7$\rightarrow$4 & 8$\rightarrow$3 & 0$\rightarrow$2 \\ \midrule
        
        ~ & SRC & 81.01 $\pm$ 0.43  &45.55 $\pm$ 0.22  &93.47 $\pm$ 0.93  &50.24 $\pm$ 7.84  &81.17 $\pm$ 0.61  &98.12 $\pm$ 0.66  &36.8 $\pm$ 8.29  &92.87 $\pm$ 0.34  &96.69 $\pm$ 0.73  &97.23 $\pm$ 0.34  \\
        \multirow{2}{*}{DDC} & DEV  & 58.37 $\pm$ 6.52  &52.47 $\pm$ 6.98  &68.38 $\pm$ 12.32  &44.83 $\pm$ 0.8  &79.85 $\pm$ 3.37  &81.3 $\pm$ 7.86  &30.37 $\pm$ 3.98  &76.23 $\pm$ 4.24  &88.77 $\pm$ 1.54  &70.08 $\pm$ 1.82  \\ 
        ~ & FST & 64.54 $\pm$ 3.09  &58.98 $\pm$ 4.79  &79.58 $\pm$ 16.53  &46.66 $\pm$ 2.93  &78.82 $\pm$ 0.63  &82.35 $\pm$ 4.15  &26.57 $\pm$ 0.32  &86.62 $\pm$ 1.7  &83.5 $\pm$ 5.43  &71.47 $\pm$ 3.25  \\
        ~ & TGT & 64.79 $\pm$ 3.03  &59.84 $\pm$ 3.51  &79.63 $\pm$ 16.41  &47.24 $\pm$ 2.33  &78.55 $\pm$ 1.31  &83.22 $\pm$ 2.6  &26.8 $\pm$ 0.28  &86.53 $\pm$ 1.31  &84.07 $\pm$ 3.96  &72.24 $\pm$ 2.18  \\ \midrule

        ~ & SRC  & 66.73 $\pm$ 1.65  &61.59 $\pm$ 2.9  &88.39 $\pm$ 3.18  &45.78 $\pm$ 3.27  &78.2 $\pm$ 1.04  &85.35 $\pm$ 1.58  &29.03 $\pm$ 0.54  &86.16 $\pm$ 0.96  &85.64 $\pm$ 3.55  &72.98 $\pm$ 1.92 \\
        \multirow{2}{*}{DCORAL} & DEV  & 58.0 $\pm$ 6.39  &55.87 $\pm$ 8.35  &83.01 $\pm$ 11.62  &40.47 $\pm$ 1.87  &78.13 $\pm$ 3.97  &73.83 $\pm$ 2.17  &27.3 $\pm$ 7.08  &84.05 $\pm$ 1.65  &77.81 $\pm$ 14.38  &73.11 $\pm$ 2.85 \\
        ~ & FST  & 69.75 $\pm$ 1.7  &60.39 $\pm$ 0.4  &89.08 $\pm$ 2.76  &43.47 $\pm$ 2.93  &78.8 $\pm$ 0.72  &87.58 $\pm$ 3.2  &29.65 $\pm$ 1.33  &86.87 $\pm$ 1.79  &86.69 $\pm$ 2.67  &85.24 $\pm$ 10.11  \\
        ~ & TGT  & 69.75 $\pm$ 1.7  &60.39 $\pm$ 0.4  &89.08 $\pm$ 2.76  &43.47 $\pm$ 2.93  &78.8 $\pm$ 0.72  &87.58 $\pm$ 3.2  &29.65 $\pm$ 1.33  &86.87 $\pm$ 1.79  &86.69 $\pm$ 2.67  &85.24 $\pm$ 10.11  \\ \midrule
        
        ~ & SRC & 66.35 $\pm$ 1.82  &62.05 $\pm$ 2.99  &88.04 $\pm$ 3.05  &42.12 $\pm$ 3.27  &78.12 $\pm$ 1.54  &85.74 $\pm$ 2.1  &28.06 $\pm$ 1.03  &85.89 $\pm$ 1.02  &85.77 $\pm$ 3.08  &77.86 $\pm$ 7.96  \\ 
        \multirow{2}{*}{HoMM} & DEV & 58.95 $\pm$ 8.77  &59.32 $\pm$ 5.43  &81.98 $\pm$ 10.78  &43.79 $\pm$ 1.12  &79.35 $\pm$ 3.02  &78.37 $\pm$ 1.84  &26.73 $\pm$ 1.68  &84.84 $\pm$ 3.12  &84.39 $\pm$ 5.49  &79.07 $\pm$ 12.67  \\ 
        ~ & FST & 70.54 $\pm$ 3.09  &58.61 $\pm$ 6.08  &89.8 $\pm$ 1.59  &41.78 $\pm$ 2.57  &79.84 $\pm$ 1.04  &92.52 $\pm$ 2.24  &29.68 $\pm$ 6.14  &88.12 $\pm$ 1.79  &90.03 $\pm$ 2.69  &95.94 $\pm$ 0.3  \\ 
        ~ & TGT & 70.54 $\pm$ 3.09  &58.61 $\pm$ 6.08  &89.8 $\pm$ 1.59  &41.78 $\pm$ 2.57  &79.84 $\pm$ 1.04  &92.52 $\pm$ 2.24  &29.68 $\pm$ 6.14  &88.12 $\pm$ 1.79  &90.03 $\pm$ 2.69  &95.94 $\pm$ 0.3  \\ \midrule

        ~ & SRC & 67.99 $\pm$ 1.31  &64.56 $\pm$ 1.7  &72.45 $\pm$ 4.17  &42.57 $\pm$ 0.58  &79.19 $\pm$ 1.76  &77.96 $\pm$ 11.18  &32.24 $\pm$ 6.6  &75.55 $\pm$ 11.49  &84.38 $\pm$ 3.64  &72.52 $\pm$ 2.0  \\
        \multirow{2}{*}{MMDA} & DEV & 74.85 $\pm$ 0.23  &57.36 $\pm$ 5.39  &77.79 $\pm$ 6.96  &50.93 $\pm$ 3.46  &79.12 $\pm$ 8.6  &78.48 $\pm$ 5.61  &34.47 $\pm$ 13.47  &77.78 $\pm$ 4.98  &82.28 $\pm$ 4.12  &80.14 $\pm$ 7.54 \\
        ~ & FST & 75.6 $\pm$ 0.44  &56.44 $\pm$ 2.68  &84.84 $\pm$ 0.57  &52.86 $\pm$ 4.57  &87.3 $\pm$ 6.65  &89.85 $\pm$ 4.64  &38.22 $\pm$ 11.16  &89.92 $\pm$ 1.95  &86.82 $\pm$ 3.99  &89.16 $\pm$ 11.66  \\
        ~ & TGT  & 74.65 $\pm$ 1.08  &43.95 $\pm$ 1.34  &89.76 $\pm$ 0.17  &53.53 $\pm$ 12.89  &95.33 $\pm$ 2.8  &94.04 $\pm$ 3.05  &39.85 $\pm$ 2.38  &91.9 $\pm$ 3.13  &92.96 $\pm$ 0.55  &93.32 $\pm$ 0.69  \\ \midrule

        ~ & SRC  & 71.15 $\pm$ 3.25  &52.69 $\pm$ 6.71  &91.42 $\pm$ 0.29  &51.04 $\pm$ 5.7  &78.51 $\pm$ 0.41  &97.1 $\pm$ 1.08  &35.83 $\pm$ 3.9  &90.74 $\pm$ 0.82  &91.66 $\pm$ 0.41  &96.5 $\pm$ 0.2  \\
        \multirow{2}{*}{DSAN} & DEV  & 62.29 $\pm$ 0.87  &55.68 $\pm$ 9.41  &92.98 $\pm$ 0.85  &50.56 $\pm$ 8.98  &91.01 $\pm$ 12.2  &97.53 $\pm$ 0.82  &35.0 $\pm$ 10.04  &90.17 $\pm$ 3.97  &95.22 $\pm$ 0.8  &96.3 $\pm$ 0.74  \\
        ~ & FST & 72.29 $\pm$ 10.88  &58.02 $\pm$ 5.95  &93.06 $\pm$ 0.07  &51.01 $\pm$ 9.68  &97.99 $\pm$ 0.38  &98.22 $\pm$ 0.4  &30.65 $\pm$ 10.67  &94.93 $\pm$ 0.52  &96.9 $\pm$ 0.7  &86.73 $\pm$ 18.88  \\
        ~ & TGT  & 80.3 $\pm$ 0.96  &62.41 $\pm$ 0.87  &93.61 $\pm$ 0.15  &51.2 $\pm$ 9.77  &97.71 $\pm$ 0.73  &98.05 $\pm$ 0.7  &31.57 $\pm$ 0.62  &94.81 $\pm$ 0.94  &96.42 $\pm$ 0.76  &86.62 $\pm$ 18.68    \\ \midrule

        ~ & SRC  & 72.62 $\pm$ 10.72  &57.07 $\pm$ 2.64  &90.21 $\pm$ 3.26  &53.12 $\pm$ 5.88  &79.31 $\pm$ 2.71  &93.02 $\pm$ 2.9  &38.4 $\pm$ 2.24  &90.77 $\pm$ 1.32  &90.63 $\pm$ 5.14  &95.92 $\pm$ 0.89  \\
        \multirow{2}{*}{DANN}& DEV  & 48.23 $\pm$ 14.38  &34.69 $\pm$ 22.28  &84.79 $\pm$ 3.45  &36.67 $\pm$ 7.9  &69.74 $\pm$ 10.49  &76.73 $\pm$ 21.63  &24.73 $\pm$ 6.19  &85.42 $\pm$ 4.72  &81.78 $\pm$ 7.34  &74.64 $\pm$ 8.48  \\
        ~ & FST & 73.29 $\pm$ 8.41  &52.63 $\pm$ 6.39  &90.96 $\pm$ 2.97  &50.92 $\pm$ 7.85  &77.03 $\pm$ 11.55  &92.61 $\pm$ 4.47  &28.65 $\pm$ 6.03  &85.19 $\pm$ 5.3  &89.84 $\pm$ 1.76  &94.51 $\pm$ 3.02  \\
        ~ & TGT & 79.97 $\pm$ 0.92  &45.49 $\pm$ 0.69  &93.1 $\pm$ 1.48  &48.65 $\pm$ 8.0  &97.7 $\pm$ 0.52  &97.08 $\pm$ 0.77  &31.73 $\pm$ 6.48  &92.36 $\pm$ 1.75  &95.97 $\pm$ 0.32  &96.81 $\pm$ 0.11   \\  \midrule
        
        ~ & SRC  & 81.01 $\pm$ 0.43  &45.55 $\pm$ 0.22  &93.47 $\pm$ 0.93  &50.24 $\pm$ 7.84  &81.17 $\pm$ 0.61  &98.12 $\pm$ 0.66  &36.8 $\pm$ 8.29  &92.87 $\pm$ 0.34  &96.69 $\pm$ 0.73  &97.23 $\pm$ 0.34  \\ 
        \multirow{2}{*}{CDAN} & DEV & 77.31 $\pm$ 4.7  &46.04 $\pm$ 0.36  &92.62 $\pm$ 0.25  &50.36 $\pm$ 6.34  &90.69 $\pm$ 11.88  &89.56 $\pm$ 13.39  &38.45 $\pm$ 4.61  &91.72 $\pm$ 0.11  &95.43 $\pm$ 1.09  &96.95 $\pm$ 0.66  \\ 
        ~ & FST  & 80.21 $\pm$ 0.85  &48.96 $\pm$ 8.04  &93.09 $\pm$ 1.55  &52.28 $\pm$ 6.79  &91.61 $\pm$ 11.06  &87.4 $\pm$ 12.86  &34.0 $\pm$ 1.86  &91.82 $\pm$ 1.54  &94.33 $\pm$ 1.38  &86.92 $\pm$ 18.15  \\ 
        ~ & TGT  & 80.49 $\pm$ 1.29  &45.65 $\pm$ 0.4  &93.71 $\pm$ 0.4  &57.68 $\pm$ 1.38  &97.99 $\pm$ 0.25  &97.94 $\pm$ 0.48  &32.33 $\pm$ 3.97  &92.33 $\pm$ 0.21  &95.36 $\pm$ 1.17  &97.24 $\pm$ 0.36  \\ \midrule

        ~ & SRC  & 80.46 $\pm$ 0.94  &47.08 $\pm$ 0.42  &94.07 $\pm$ 0.82  &52.8 $\pm$ 9.6  &87.02 $\pm$ 10.62  &98.21 $\pm$ 0.12  &35.64 $\pm$ 2.5  &96.05 $\pm$ 0.88  &97.32 $\pm$ 0.48  &96.93 $\pm$ 0.04  \\
        \multirow{2}{*}{DIRT-T} & DEV  & 81.05 $\pm$ 0.6  &51.77 $\pm$ 9.58  &93.43 $\pm$ 0.31  &63.65 $\pm$ 0.58  &97.78 $\pm$ 2.24  &96.79 $\pm$ 0.66  &28.47 $\pm$ 12.7  &97.48 $\pm$ 0.68  &96.82 $\pm$ 0.47  &96.18 $\pm$ 0.29 \\
        ~ & FST & 77.55 $\pm$ 2.03  &61.69 $\pm$ 0.21  &94.75 $\pm$ 0.62  &55.03 $\pm$ 9.14  &95.15 $\pm$ 0.09  &96.16 $\pm$ 0.1  &31.07 $\pm$ 4.55  &97.14 $\pm$ 0.45  &94.57 $\pm$ 1.15  &95.84 $\pm$ 0.38  \\
        ~ & TGT & 76.54 $\pm$ 7.66  &58.13 $\pm$ 9.8  &94.2 $\pm$ 0.75  &64.07 $\pm$ 0.99  &98.88 $\pm$ 0.35  &97.02 $\pm$ 0.57  &25.48 $\pm$ 1.38  &97.75 $\pm$ 1.04  &97.27 $\pm$ 0.23  &95.34 $\pm$ 0.15  \\ \midrule

        ~ & SRC & 72.67 $\pm$ 9.46  &44.68 $\pm$ 1.89  &90.4 $\pm$ 2.83  &42.97 $\pm$ 2.37  &85.14 $\pm$ 8.14  &94.96 $\pm$ 0.78  &27.09 $\pm$ 6.34  &91.42 $\pm$ 1.34  &93.05 $\pm$ 1.58  &96.52 $\pm$ 0.34  \\ 
        \multirow{2}{*}{CoDATS} & DEV & 71.0 $\pm$ 6.85  &46.65 $\pm$ 7.26  &90.04 $\pm$ 1.38  &49.79 $\pm$ 5.35  &96.7 $\pm$ 1.14  &85.28 $\pm$ 10.1  &26.62 $\pm$ 5.1  &83.73 $\pm$ 8.12  &90.33 $\pm$ 2.95  &85.88 $\pm$ 17.74   \\ 
        ~ & FST & 67.12 $\pm$ 6.19  &41.55 $\pm$ 3.23  &86.37 $\pm$ 9.53  &48.07 $\pm$ 9.42  &92.02 $\pm$ 0.77  &87.57 $\pm$ 4.06  &30.83 $\pm$ 11.48  &85.02 $\pm$ 4.35  &91.2 $\pm$ 2.95  &83.05 $\pm$ 4.09  \\ 
        ~ & TGT  & 71.88 $\pm$ 11.92  &45.43 $\pm$ 0.62  &92.74 $\pm$ 0.84  &50.04 $\pm$ 4.85  &88.62 $\pm$ 10.25  &94.14 $\pm$ 2.5  &39.41 $\pm$ 4.96  &92.41 $\pm$ 1.27  &93.18 $\pm$ 0.78  &96.24 $\pm$ 0.39  
        \\ \midrule

        ~ & SRC & 67.98 $\pm$ 6.26  &54.26 $\pm$ 9.26  &73.39 $\pm$ 2.19  &41.77 $\pm$ 4.51  &76.82 $\pm$ 1.51  &78.86 $\pm$ 6.94  &28.85 $\pm$ 2.36  &75.54 $\pm$ 8.54  &83.2 $\pm$ 6.0  &72.54 $\pm$ 2.24  \\ 
        \multirow{2}{*}{AdvSKM} & DEV & 54.69 $\pm$ 8.05  &50.05 $\pm$ 10.74  &72.87 $\pm$ 12.52  &41.82 $\pm$ 11.42  &78.45 $\pm$ 2.21  &83.95 $\pm$ 3.82  &31.0 $\pm$ 7.3  &86.74 $\pm$ 2.26  &84.56 $\pm$ 6.23  &67.86 $\pm$ 3.61 \\     
        ~ & FST  & 68.96 $\pm$ 4.13  &54.2 $\pm$ 8.79  &64.87 $\pm$ 4.03  &41.11 $\pm$ 6.5  &78.07 $\pm$ 1.86  &81.77 $\pm$ 4.81  &25.07 $\pm$ 2.01  &79.39 $\pm$ 5.31  &81.71 $\pm$ 5.48  &70.96 $\pm$ 3.53  \\ 
        ~ & TGT  & 62.98 $\pm$ 2.18  &54.78 $\pm$ 9.01  &70.04 $\pm$ 12.04  &42.16 $\pm$ 3.71  &77.13 $\pm$ 3.71  &82.13 $\pm$ 4.29  &34.45 $\pm$ 6.9  &82.13 $\pm$ 11.77  &83.66 $\pm$ 5.29  &69.3 $\pm$ 5.07  \\ \midrule

        ~ & SRC & 60.77 $\pm$ 3.56  &57.35 $\pm$ 9.92  &91.62 $\pm$ 0.56  &43.61 $\pm$ 2.41  &95.69 $\pm$ 0.62  &90.77 $\pm$ 0.8  &32.39 $\pm$ 2.81  &91.3 $\pm$ 3.6  &85.5 $\pm$ 3.42  &92.7 $\pm$ 3.73  \\ 
        \multirow{2}{*}{SASA} & DEV & 62.7 $\pm$ 5.45  &46.35 $\pm$ 4.49  &89.28 $\pm$ 2.07  &43.57 $\pm$ 3.87  &90.01 $\pm$ 9.51  &91.4 $\pm$ 4.12  &28.6 $\pm$ 2.22  &92.9 $\pm$ 0.21  &86.81 $\pm$ 7.2  &90.35 $\pm$ 9.83   \\ 
        ~ & FST & 67.31 $\pm$ 3.84  &57.45 $\pm$ 5.96  &91.68 $\pm$ 0.46  &45.82 $\pm$ 2.09  &92.49 $\pm$ 3.12  &95.49 $\pm$ 0.92  &28.44 $\pm$ 4.26  &91.6 $\pm$ 1.35  &90.07 $\pm$ 0.58  &96.94 $\pm$ 0.24  \\ 
        ~ & TGT & 65.38 $\pm$ 0.81  &57.56 $\pm$ 6.8  &91.63 $\pm$ 0.85  &44.38 $\pm$ 3.25  &94.7 $\pm$ 0.33  &96.44 $\pm$ 0.83  &28.24 $\pm$ 0.79  &93.17 $\pm$ 0.86  &89.64 $\pm$ 1.17  &96.47 $\pm$ 0.45   \\ \bottomrule
    \end{NiceTabular}}
\end{table}

\begin{table}[!ht]
    \centering
    \caption{Detailed results of 10 scenarios on MFD dataset in terms of MF1 score.}
    \label{tbl:detailed_MFD}
    \resizebox{\textwidth}{!}{
    \begin{NiceTabular}{l|c|c|c|c|c|c|c|c|c|c|c} 
    \toprule
    Algorithm &RISK & 0$\rightarrow$1 & 0$\rightarrow$3 & 1$\rightarrow$0 & 1$\rightarrow$2 & 1$\rightarrow$3 & 2$\rightarrow$1 & 2$\rightarrow$3 & 3$\rightarrow$0 & 3$\rightarrow$1 & 3$\rightarrow$2 \\ \midrule

    ~ & SRC & 77.75 $\pm$ 3.65  &80.57 $\pm$ 5.68  &40.64 $\pm$ 3.95  &82.68 $\pm$ 3.67  &98.88 $\pm$ 1.87  &96.47 $\pm$ 1.17  &96.69 $\pm$ 0.83  &39.73 $\pm$ 1.44  &99.97 $\pm$ 0.05  &83.54 $\pm$ 7.12  \\
    ~\multirow{2}{*}{DDC} & DEV & 73.23 $\pm$ 2.22  &79.5 $\pm$ 1.1  &44.34 $\pm$ 2.68  &84.9 $\pm$ 0.2  &99.92 $\pm$ 0.08  &96.8 $\pm$ 0.21  &96.89 $\pm$ 0.71  &40.69 $\pm$ 1.26  &99.92 $\pm$ 0.14  &85.93 $\pm$ 4.49  \\
    ~ & FST & 77.75 $\pm$ 3.65  &80.57 $\pm$ 5.68  &40.64 $\pm$ 3.95  &82.68 $\pm$ 3.67  &98.88 $\pm$ 1.87  &96.47 $\pm$ 1.17  &96.69 $\pm$ 0.83  &39.73 $\pm$ 1.44  &99.97 $\pm$ 0.05  &83.54 $\pm$ 7.12  \\
    ~ & TGT & 74.5 $\pm$ 5.56  &79.31 $\pm$ 6.2  &48.91 $\pm$ 6.24  &89.34 $\pm$ 2.16  &99.89 $\pm$ 0.19  &96.21 $\pm$ 3.04  &96.34 $\pm$ 3.07  &46.59 $\pm$ 5.45  &100.0 $\pm$ 0.0  &84.3 $\pm$ 1.83  \\\midrule
    
    ~ & SRC & 76.25 $\pm$ 2.91  &81.06 $\pm$ 2.83  &38.33 $\pm$ 2.52  &83.74 $\pm$ 4.06  &99.97 $\pm$ 0.05  &97.81 $\pm$ 1.17  &98.23 $\pm$ 0.42  &42.77 $\pm$ 4.71  &99.97 $\pm$ 0.05  &81.79 $\pm$ 4.29  \\
    ~\multirow{2}{*}{DCORAL} & DEV & 58.62 $\pm$ 4.51  &80.91 $\pm$ 4.14  &47.95 $\pm$ 4.74  &83.02 $\pm$ 2.71  &99.97 $\pm$ 0.05  &96.43 $\pm$ 3.14  &95.9 $\pm$ 3.59  &43.84 $\pm$ 5.29  &97.26 $\pm$ 4.32  &81.89 $\pm$ 2.88  \\
    ~ & FST & 76.25 $\pm$ 2.91  &81.06 $\pm$ 2.83  &38.33 $\pm$ 2.52  &83.74 $\pm$ 4.06  &99.97 $\pm$ 0.05  &97.81 $\pm$ 1.17  &98.23 $\pm$ 0.42  &42.77 $\pm$ 4.71  &99.97 $\pm$ 0.05  &81.79 $\pm$ 4.29  \\
    ~ & TGT & 79.03 $\pm$ 8.83  &81.83 $\pm$ 6.29  &40.83 $\pm$ 5.01  &82.71 $\pm$ 0.76  &100.0 $\pm$ 0.0  &97.01 $\pm$ 2.4  &98.01 $\pm$ 0.67  &47.13 $\pm$ 7.82  &97.73 $\pm$ 3.93  &83.72 $\pm$ 3.04  \\\midrule

    ~ & SRC & 79.51 $\pm$ 2.86  &81.43 $\pm$ 3.98  &42.62 $\pm$ 3.42  &84.52 $\pm$ 2.5  &100.0 $\pm$ 0.0  &96.33 $\pm$ 0.91  &96.83 $\pm$ 0.67  &42.47 $\pm$ 5.16  &100.0 $\pm$ 0.0  &80.9 $\pm$ 3.81  \\
    ~\multirow{2}{*}{HoMM} & DEV & 68.19 $\pm$ 8.79  &74.69 $\pm$ 3.09  &50.02 $\pm$ 2.47  &81.95 $\pm$ 4.59  &100.0 $\pm$ 0.0  &98.72 $\pm$ 0.91  &96.32 $\pm$ 4.21  &44.37 $\pm$ 6.4  &100.0 $\pm$ 0.0  &83.53 $\pm$ 2.07  \\
    ~ & FST & 79.55 $\pm$ 4.46  &81.33 $\pm$ 5.44  &41.95 $\pm$ 3.48  &83.93 $\pm$ 5.12  &100.0 $\pm$ 0.0  &96.96 $\pm$ 1.5  &96.66 $\pm$ 1.47  &40.33 $\pm$ 1.0  &99.73 $\pm$ 0.47  &81.23 $\pm$ 4.97  \\
    ~ & TGT & 80.8 $\pm$ 2.46  &84.77 $\pm$ 0.41  &42.31 $\pm$ 5.9  &84.28 $\pm$ 1.32  &100.0 $\pm$ 0.0  &97.06 $\pm$ 3.25  &98.61 $\pm$ 0.08  &43.14 $\pm$ 5.23  &96.28 $\pm$ 6.45  &84.6 $\pm$ 2.99  \\\midrule
    
    ~ & SRC & 83.65 $\pm$ 1.56  &85.29 $\pm$ 0.39  &44.25 $\pm$ 1.11  &93.36 $\pm$ 1.63  &100.0 $\pm$ 0.0  &99.43 $\pm$ 0.99  &100.0 $\pm$ 0.0  &42.43 $\pm$ 0.7  &99.97 $\pm$ 0.05  &93.58 $\pm$ 0.98  \\
    ~\multirow{2}{*}{MMDA} & DEV & 81.84 $\pm$ 2.46  &77.73 $\pm$ 4.64  &39.48 $\pm$ 1.89  &92.62 $\pm$ 1.06  &100.0 $\pm$ 0.0  &99.73 $\pm$ 0.47  &99.75 $\pm$ 0.43  &40.37 $\pm$ 0.64  &100.0 $\pm$ 0.0  &93.0 $\pm$ 2.45  \\
    ~ & FST & 83.65 $\pm$ 1.56  &85.29 $\pm$ 0.39  &44.25 $\pm$ 1.11  &93.36 $\pm$ 1.63  &100.0 $\pm$ 0.0  &99.43 $\pm$ 0.99  &100.0 $\pm$ 0.0  &42.43 $\pm$ 0.7  &99.97 $\pm$ 0.05  &93.58 $\pm$ 0.98  \\
    ~ & TGT & 82.44 $\pm$ 4.47  &85.55 $\pm$ 0.14  &49.35 $\pm$ 5.02  &94.07 $\pm$ 2.72  &100.0 $\pm$ 0.0  &99.75 $\pm$ 0.43  &100.0 $\pm$ 0.0  &48.36 $\pm$ 6.0  &100.0 $\pm$ 0.0  &94.84 $\pm$ 2.15  \\\midrule

    ~ & SRC & 49.4 $\pm$ 12.04  &53.18 $\pm$ 15.55  &22.92 $\pm$ 4.56  &58.18 $\pm$ 19.64  &63.36 $\pm$ 8.86  &73.34 $\pm$ 13.67  &66.5 $\pm$ 8.46  &20.34 $\pm$ 0.34  &66.24 $\pm$ 3.24  &52.74 $\pm$ 15.14  \\
    ~\multirow{2}{*}{DSAN} & DEV & 80.0 $\pm$ 4.25  &82.22 $\pm$ 3.08  &39.25 $\pm$ 1.13  &80.2 $\pm$ 2.78  &99.54 $\pm$ 0.37  &97.51 $\pm$ 0.83  &98.14 $\pm$ 0.05  &39.37 $\pm$ 0.24  &99.97 $\pm$ 0.05  &84.94 $\pm$ 2.55  \\
    ~ & FST & 79.94 $\pm$ 9.63  &74.39 $\pm$ 9.47  &41.18 $\pm$ 0.27  &69.86 $\pm$ 1.64  &99.78 $\pm$ 0.38  &89.27 $\pm$ 8.72  &83.4 $\pm$ 1.1  &41.43 $\pm$ 0.21  &100.0 $\pm$ 0.0  &68.91 $\pm$ 2.45  \\
    ~ & TGT & 80.03 $\pm$ 3.7  &83.63 $\pm$ 1.04  &41.31 $\pm$ 0.05  &88.01 $\pm$ 1.57  &100.0 $\pm$ 0.0  &96.63 $\pm$ 2.3  &98.0 $\pm$ 1.69  &41.25 $\pm$ 0.24  &99.54 $\pm$ 0.8  &88.1 $\pm$ 0.68  \\\midrule

    ~ & SRC & 79.08 $\pm$ 11.0  &77.0 $\pm$ 12.47  &42.26 $\pm$ 1.86  &84.0 $\pm$ 0.75  &100.0 $\pm$ 0.0  &99.86 $\pm$ 0.17  &98.93 $\pm$ 1.57  &44.25 $\pm$ 6.3  &99.04 $\pm$ 1.66  & 84.69 $\pm$ 0.76  \\
    ~\multirow{2}{*}{DANN} & DEV & 85.66 $\pm$ 0.05  &82.42 $\pm$ 4.98  &49.5 $\pm$ 4.33  &82.05 $\pm$ 2.99  &99.89 $\pm$ 0.19  &96.41 $\pm$ 1.86  &97.35 $\pm$ 1.26  &45.07 $\pm$ 6.37  &98.88 $\pm$ 1.73  &82.32 $\pm$ 3.13  \\
    ~ & FST & 79.08 $\pm$ 11.0  &77.0 $\pm$ 12.47  &42.26 $\pm$ 1.86  &84.0 $\pm$ 0.75  &100.0 $\pm$ 0.0  &99.86 $\pm$ 0.17  &98.93 $\pm$ 1.57  &44.25 $\pm$ 6.3  &99.04 $\pm$ 1.66  &84.69 $\pm$ 0.76  \\
    ~ & TGT & 83.44 $\pm$ 1.72  &83.99 $\pm$ 1.5  &51.52 $\pm$ 0.38  &84.19 $\pm$ 2.1  &100.0 $\pm$ 0.0  &99.95 $\pm$ 0.05  &99.95 $\pm$ 0.09  &51.54 $\pm$ 0.45  &100.0 $\pm$ 0.0  &86.04 $\pm$ 0.88  \\ \midrule
    
    ~ & SRC & 51.95 $\pm$ 21.77  &64.23 $\pm$ 13.07  &45.8 $\pm$ 7.81  &74.42 $\pm$ 21.95  &99.97 $\pm$ 0.05  &87.62 $\pm$ 4.76  &87.85 $\pm$ 5.56  &41.24 $\pm$ 0.13  &100.0 $\pm$ 0.0  &71.27 $\pm$ 21.39  \\
    ~\multirow{2}{*}{CDAN} & DEV & 60.62 $\pm$ 19.24  &53.28 $\pm$ 6.08  &44.64 $\pm$ 6.73  &92.71 $\pm$ 0.0  &99.92 $\pm$ 0.08  &99.97 $\pm$ 0.05  &100.0 $\pm$ 0.0  &44.85 $\pm$ 6.52  &99.97 $\pm$ 0.05  &87.46 $\pm$ 7.91  \\
    ~ & FST & 43.76 $\pm$ 24.81  &48.76 $\pm$ 7.21  &58.94 $\pm$ 18.99  &59.14 $\pm$ 9.28  &100.0 $\pm$ 0.0  &63.03 $\pm$ 20.75  &62.88 $\pm$ 23.22  &58.09 $\pm$ 19.55  &100.0 $\pm$ 0.0  &54.3 $\pm$ 2.37  \\
    ~ & TGT & 84.97 $\pm$ 0.62  &85.52 $\pm$ 0.12  &52.39 $\pm$ 0.49  &85.96 $\pm$ 0.9  &100.0 $\pm$ 0.0  &99.62 $\pm$ 0.59  &99.7 $\pm$ 0.45  &51.64 $\pm$ 1.03  &100.0 $\pm$ 0.0  &86.64 $\pm$ 2.12  \\\midrule
    
    ~ & SRC & 88.22 $\pm$ 3.83  &90.47 $\pm$ 3.95  &77.4 $\pm$ 20.28  &92.79 $\pm$ 0.0  &100.0 $\pm$ 0.0  &94.2 $\pm$ 2.88  &99.89 $\pm$ 0.09  &78.34 $\pm$ 0.64  &99.21 $\pm$ 1.3  &92.55 $\pm$ 0.42  \\
    ~\multirow{2}{*}{DIRT-T} & DEV & 85.84 $\pm$ 2.05  &87.08 $\pm$ 2.05  &60.73 $\pm$ 11.9  &92.79 $\pm$ 0.0  &100.0 $\pm$ 0.0  &100.0 $\pm$ 0.0  &99.92 $\pm$ 0.14  &54.52 $\pm$ 2.64  &100.0 $\pm$ 0.0  &92.79 $\pm$ 0.0  \\
    ~ & FST & 88.22 $\pm$ 3.83  &90.47 $\pm$ 3.95  &77.4 $\pm$ 20.28  &92.79 $\pm$ 0.0  &100.0 $\pm$ 0.0  &94.2 $\pm$ 2.88  &99.89 $\pm$ 0.09  &78.34 $\pm$ 0.64  &99.21 $\pm$ 1.3  &92.55 $\pm$ 0.42  \\
    ~ & TGT & 88.94 $\pm$ 3.28  &89.03 $\pm$ 3.1  &79.92 $\pm$ 8.53  &92.79 $\pm$ 0.0  &100.0 $\pm$ 0.0  &99.97 $\pm$ 0.05  &99.97 $\pm$ 0.05  &84.71 $\pm$ 4.61  &100.0 $\pm$ 0.0  &92.79 $\pm$ 0.0  \\\midrule

    ~ & SRC & 68.07 $\pm$ 16.68  &71.34 $\pm$ 15.06  &50.24 $\pm$ 1.41  &77.97 $\pm$ 10.31  &99.84 $\pm$ 0.14  &87.22 $\pm$ 20.73  &93.55 $\pm$ 6.3  &39.92 $\pm$ 19.07  &99.24 $\pm$ 0.93  &82.37 $\pm$ 1.64  \\
    ~\multirow{2}{*}{CoDATS} & DEV & 67.07 $\pm$ 8.92  &70.44 $\pm$ 17.28  &46.92 $\pm$ 3.84  &84.46 $\pm$ 5.92  &100.0 $\pm$ 0.0  &98.25 $\pm$ 0.62  &99.95 $\pm$ 0.09  &50.86 $\pm$ 6.06  &99.97 $\pm$ 0.05  &85.0 $\pm$ 2.71  \\
    ~ & FST & 73.07 $\pm$ 17.08  &71.38 $\pm$ 20.01  &50.28 $\pm$ 1.97  &85.72 $\pm$ 0.21  &100.0 $\pm$ 0.0  &99.7 $\pm$ 0.26  &98.91 $\pm$ 1.9  &52.32 $\pm$ 4.62  &100.0 $\pm$ 0.0  &85.97 $\pm$ 2.07  \\
    ~ & TGT & 67.42 $\pm$ 13.32  &89.43 $\pm$ 5.62  &49.92 $\pm$ 13.69  &89.05 $\pm$ 4.73  &100.0 $\pm$ 0.0  &99.45 $\pm$ 0.26  &99.21 $\pm$ 0.79  &56.48 $\pm$ 18.91  &99.92 $\pm$ 0.14  &91.16 $\pm$ 4.12  \\\midrule
    
    ~ & SRC & 74.86 $\pm$ 9.06  &78.41 $\pm$ 7.26  &39.2 $\pm$ 2.65  &81.86 $\pm$ 2.3  &99.67 $\pm$ 0.57  &96.39 $\pm$ 3.72  &97.37 $\pm$ 1.57  &45.44 $\pm$ 4.83  &100.0 $\pm$ 0.0  &84.43 $\pm$ 2.17  \\
    ~\multirow{2}{*}{AdvSKM} & DEV & 75.59 $\pm$ 2.3  &80.92 $\pm$ 0.56  &48.91 $\pm$ 1.02  &84.72 $\pm$ 5.25  &99.84 $\pm$ 0.22  &95.68 $\pm$ 3.25  &95.82 $\pm$ 3.27  &48.48 $\pm$ 6.2  &98.19 $\pm$ 1.81  &83.74 $\pm$ 2.19  \\
    ~ & FST & 74.86 $\pm$ 9.06  &78.41 $\pm$ 7.26  &39.2 $\pm$ 2.65  &81.86 $\pm$ 2.3  &99.67 $\pm$ 0.57  &96.39 $\pm$ 3.72  &97.37 $\pm$ 1.57  &45.44 $\pm$ 4.83  &100.0 $\pm$ 0.0  &84.43 $\pm$ 2.17  \\
    ~ & TGT & 76.64 $\pm$ 4.82  &82.24 $\pm$ 2.9  &43.81 $\pm$ 6.29  &83.1 $\pm$ 2.19  &99.97 $\pm$ 0.05  &98.63 $\pm$ 1.03  &98.85 $\pm$ 0.93  &49.44 $\pm$ 2.33  &100.0 $\pm$ 0.0  &82.03 $\pm$ 4.25  \\\midrule
    
    ~ & SRC & 55.98 $\pm$ 6.2  &56.6 $\pm$ 3.41  &51.84 $\pm$ 0.63  &81.25 $\pm$ 2.19  &96.06 $\pm$ 2.55  &91.74 $\pm$ 3.26  &92.87 $\pm$ 2.6  &53.09 $\pm$ 1.75  &99.95 $\pm$ 0.05  &85.29 $\pm$ 3.91  \\
    ~\multirow{2}{*}{SASA} & DEV & 63.87 $\pm$ 5.49  &57.44 $\pm$ 4.39  &47.27 $\pm$ 5.2  &81.35 $\pm$ 2.04  &99.95 $\pm$ 0.05  &98.5 $\pm$ 1.82  &97.0 $\pm$ 4.07  &46.91 $\pm$ 3.23  &100.0 $\pm$ 0.0  &83.64 $\pm$ 1.16  \\
    ~ & FST & 62.61 $\pm$ 10.45  &61.12 $\pm$ 12.94  &51.68 $\pm$ 0.07  &81.19 $\pm$ 3.03  &99.04 $\pm$ 1.04  &92.7 $\pm$ 1.44  &93.93 $\pm$ 3.28  &51.63 $\pm$ 2.38  &99.95 $\pm$ 0.05  &83.54 $\pm$ 1.73  \\
    ~ & TGT & 67.45 $\pm$ 13.18  &68.85 $\pm$ 14.67  &51.07 $\pm$ 1.05  &85.25 $\pm$ 1.57  &99.65 $\pm$ 0.55  &96.73 $\pm$ 3.84  &96.98 $\pm$ 3.33  &51.33 $\pm$ 0.29  &100.0 $\pm$ 0.0  &85.39 $\pm$ 3.71  \\
    
    \bottomrule
    
    \end{NiceTabular}}
\end{table}

\end{document}